\documentclass{article}

\usepackage[final]{corl_2021} 

\usepackage{amsmath,amssymb,graphicx,color,booktabs,bm,relsize,enumitem,multirow,amsthm,epsfig, caption,mathtools,subfig,adjustbox}
\usepackage{bbm}
\usepackage{wrapfig}
\usepackage{tabularx}
\usepackage{algorithm,algorithmic}

\title{Offline-to-Online Reinforcement Learning via
Balanced Replay and Pessimistic Q-Ensemble}

\author{
  Seunghyun Lee\thanks{Equal Contribution. Correspondence to \texttt{\{seunghyun.lee, younggyo.seo\}@kaist.ac.kr}} $^{ , \dagger}$
  \quad
  Younggyo Seo\footnotemark[1] $^{ , \dagger}$
  \quad
  Kimin Lee$^{\ddagger}$
  \quad
  Pieter Abbeel$^{\ddagger}$
  \quad
  \textbf{Jinwoo Shin}$^{\dagger}$
  \\
  $^{\dagger}$ Korea Advanced Institute of Science and Technology 
  \\
  $^{\ddagger}$University of California, Berkeley 
}

\begin{document}
\maketitle


\begin{abstract}
Recent advance in deep offline reinforcement learning (RL) has made it possible to train strong robotic agents from offline datasets. 
However, depending on the quality of the trained agents and the application being considered, it is often desirable to fine-tune such agents via further online interactions.
In this paper, we observe that state-action distribution shift may lead to severe bootstrap error during fine-tuning, 
which destroys the good initial policy obtained via offline RL. 
To address this issue, we first propose a balanced replay scheme that prioritizes samples encountered online while also encouraging the use of near-on-policy samples from the offline dataset. 
Furthermore, we leverage multiple Q-functions trained pessimistically offline, thereby preventing overoptimism concerning unfamiliar actions at novel states during the initial training phase. 
We show that the proposed method improves sample-efficiency and final performance of the fine-tuned robotic agents on various locomotion and manipulation tasks.
Our code is available at: \url{https://github.com/shlee94/Off2OnRL}.
\end{abstract}


\section{Introduction}
Deep offline reinforcement learning (RL) \citep{levine2020offline} has the potential to train strong robotic agents without any further environment interaction by leveraging deep neural networks and huge offline datasets. 
Accordingly, the research community has demonstrated that offline RL can train both simulated \citep{fujimoto2019off, kumar2019stabilizing, wu2019behavior, siegel2020keep, agarwal2020optimistic, kidambi2020morel,yu2020mopo,kumar2020conservative} and real \citep{siegel2020keep,singh2020cog} robots that are often more performant than the behavior policy that generated the offline dataset.
However, thusly trained offline RL agents may be suboptimal, for (a) the dataset they were trained on may be suboptimal; and (b) environment in which they are deployed may be different from the environment in which the dataset was generated. This necessitates an online fine-tuning procedure, where the robot improves by gathering additional samples. 

Off-policy RL algorithms are well-suited for offline-to-online RL, since they can leverage both offline and online samples. 
Fine-tuning an offline RL agent using a conventional off-policy RL algorithm, however, is difficult due to distribution shift, i.e., 
the robot may encounter unfamiliar state-action regime that is not covered by the offline dataset. The Q-function cannot provide accurate value estimates for such out-of-distribution (OOD) online samples, and updates with such samples lead to severe bootstrap error. 
This leads to policy updates in an arbitrary direction, destroying the good initial policy obtained by offline RL. 

To address state-action distribution shift, 
we first introduce a balanced replay scheme that enables us to provide the robotic agent with near-on-policy samples from the offline dataset, in addition to samples gathered online. 
Specifically, we train a network that measures the \textit{online-ness} of available samples, then prioritize samples according to this measure. 
This adjusts the sampling distribution for Q-learning to be closer to online samples, which enables timely value propagation and more accurate policy evaluation in the novel state-action regime. 

However, we find that the above sampling scheme is not enough, 
for the Q-function may be overoptimistic about unseen actions at novel online states.
This misleads the robot to prefer potentially bad actions, and in turn, more severe distribution shift and bootstrap error.
We therefore propose a pessimistic Q-ensemble scheme. 
In particular, we first observe that a specific class of offline RL algorithms that train pessimistic Q-functions \citep{yu2020mopo,kumar2020conservative} make an excellent starting point for offline-to-online RL. 
When trained as such, the Q-function implicitly constrains the policy to stay near the behavior policy during the initial fine-tuning phase. 
Building on this observation, we leverage multiple pessimistic Q-functions, which guides the robotic agent with a more high-resolution pessimism and stabilizes fine-tuning. 

In our experiments, we demonstrate the strength of our method based on (1) MuJoCo \citep{todorov2012mujoco} locomotion tasks from the D4RL \citep{fu2020d4rl} benchmark suite, and (2) vision-based robotic manipulation tasks from \citet{singh2020cog}.
We show that our method achieves stable training during fine-tuning, while outperforming all baseline methods considered, both in terms of final performance and sample-efficiency.
We provide a thorough analysis of each component of our method.

\section{Background}

\label{section:background}
\noindent {\bf Reinforcement learning.} 
We consider the standard RL framework, where an agent interacts with the environment so as to maximize the expected total return. More formally, at each timestep $t$, the agent observes a state $s_{t}$, and performs an action $a_{t}$ according to its policy $\pi$. The environment rewards the agent with $r_{t}$, then transitions to the next state $s_{t+1}$. The agent's objective is to maximize the expected return $\mathbb{E}_{\pi}[\sum_{t=0}^{\infty} \gamma^{t} r_{t}]$, where $\gamma\in [0,1)$ is the discount factor. The unnormalized stationary state-action distribution under $\pi$ is defined as $d^{\pi}(s,a) := \sum_{t=0}^{\infty} \gamma^{t} d_{t}^{\pi}(s,a)$, where 
$d^{\pi}_{t}(s,a)$ denotes the state-action distribution at timestep $t$ of the Markov chain defined by the fixed policy $\pi$.

\noindent {\bf Soft actor-critic.}
We mainly consider off-policy RL algorithms, a class of algorithms that can, in principle, train an agent with samples generated by any behavior policy. 
In particular, soft actor-critic [SAC; \citealp{haarnoja2018soft}] is an off-policy actor-critic algorithm that learns a soft Q-function $Q_{\theta}(s,a)$ parameterized by $\theta$ and a stochastic policy $\pi_{\phi}$ modeled as a Gaussian, parameterized by $\phi$. SAC alternates between critic and actor updates by minimizing the following objectives, respectively:
\begin{align}
\label{eq:sac_critic}
\mathcal{L}^{\tt{SAC}}_{\tt{critic}}(\theta) &= \underset{{(s,a,s') \sim \mathcal{B}}}{\mathbb{E}}\bigg[\big(Q_{\theta}(s, a) - r(s, a) - \gamma \underset{{a'\sim \pi_{\phi}}}{\mathbb{E}}\big[ Q_{\bar{\theta}}(s',a') - \alpha \log \pi_{\phi}(a'|s') \big]    \big)^{2}\bigg], \\
\label{eq:sac_actor}
\mathcal{L}^{\tt{SAC}}_{\tt{actor}}(\phi) &= \underset{s\sim \mathcal{B}, a\sim \pi_{\phi}}{\mathbb{E}}\big[ \alpha\log\pi_{\phi}(a|s) - Q_{\theta}(s,a) \big],
\end{align}
where $\mathcal{B}$ is the replay buffer, $\bar{\theta}$ the delayed parameters, and $\alpha$ the temperature parameter. 

\begin{figure}[htb!]
\centering
\subfloat[State-action distribution shift \label{fig:dist_shift}]{%
  \includegraphics[width=0.31\textwidth]{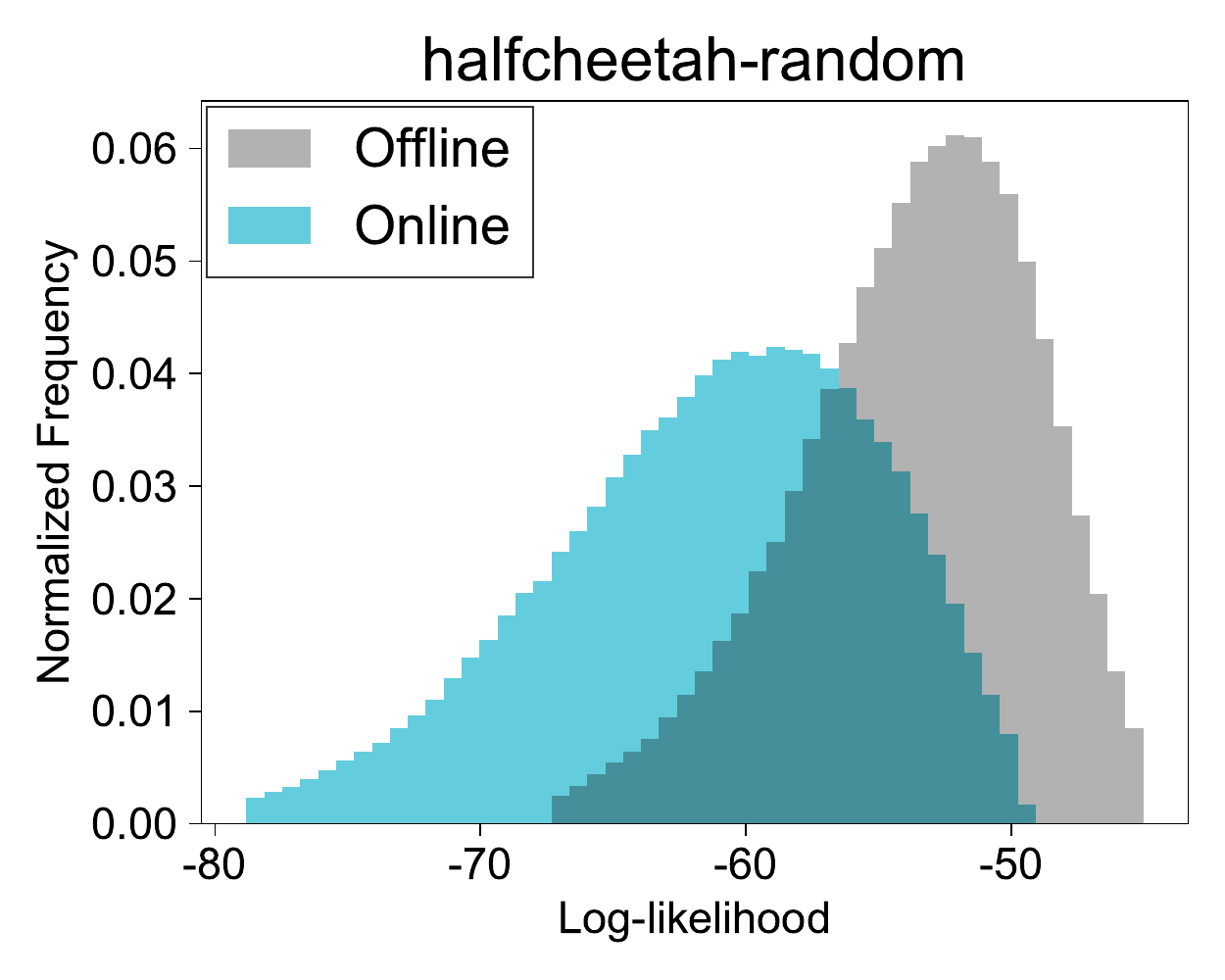}
}
\hfill
\subfloat[Sample selection \label{fig:obs_sample}]{%
  \includegraphics[width=0.31\textwidth]{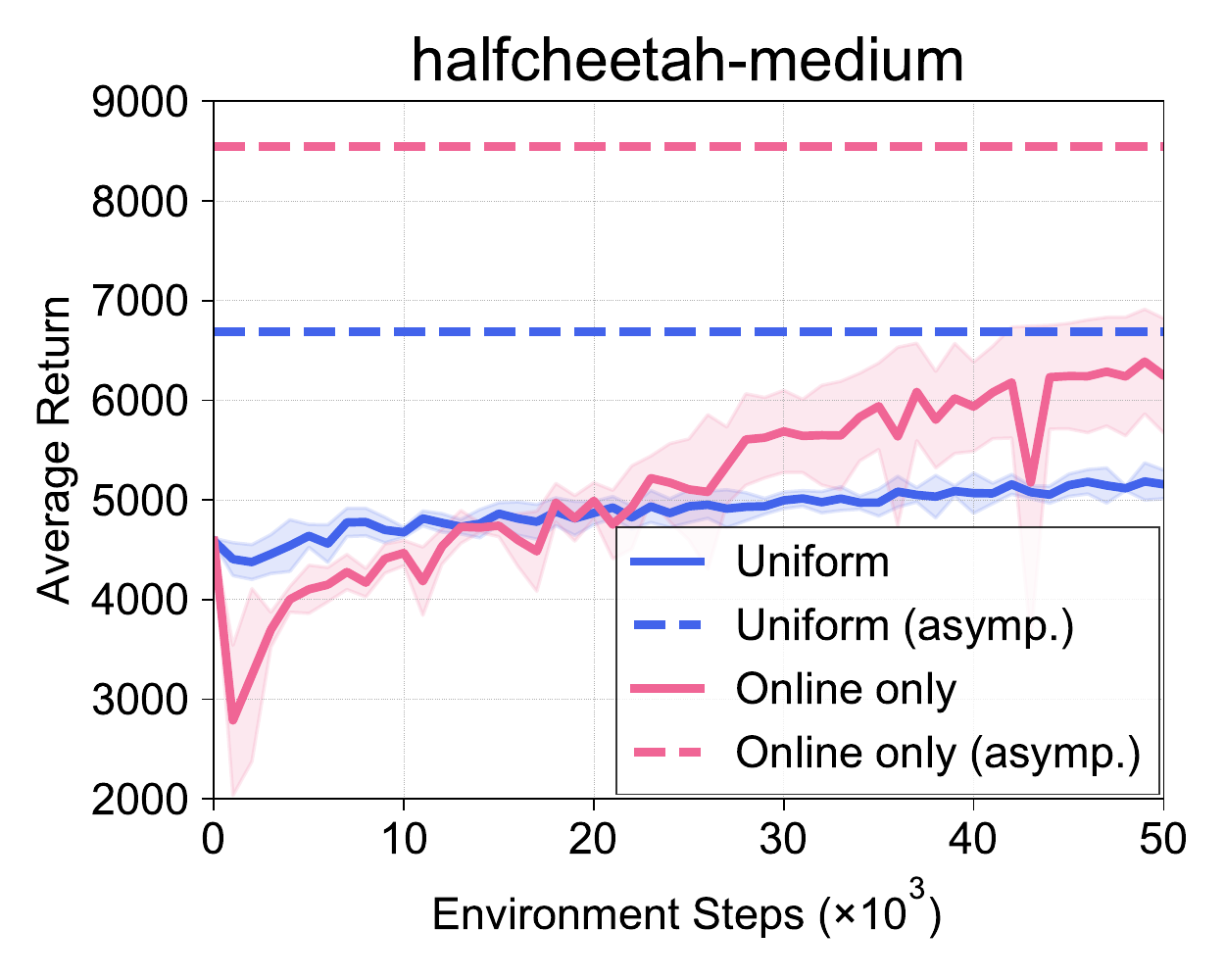}
}
\hfill
\subfloat[Choice of offline Q-function \label{fig:obs_q_init}]{%
  \includegraphics[width=0.31\textwidth]{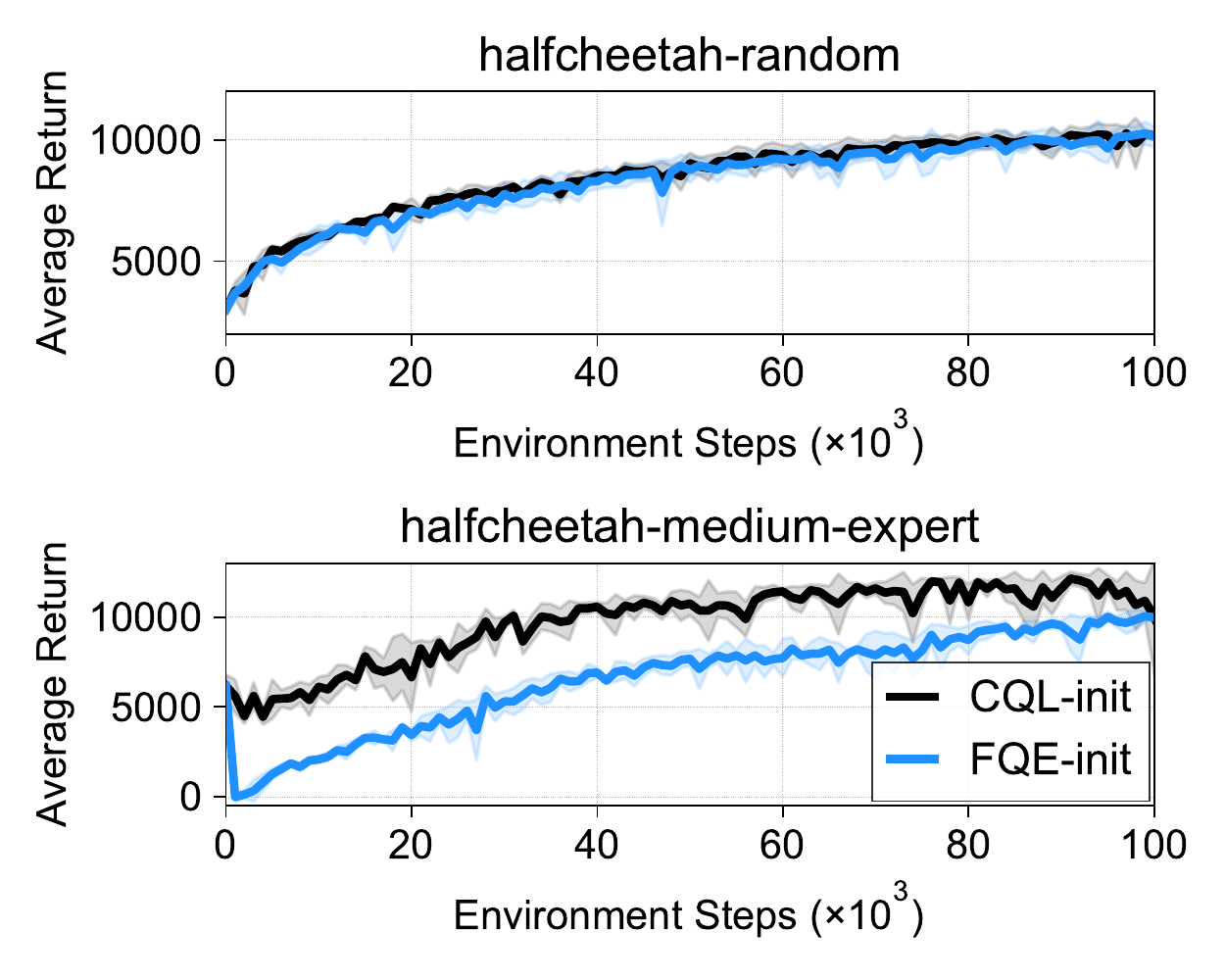}
}
\hfill
\caption{
(a) Log-likelihood estimates of (i) offline samples and (ii) online samples gathered by the offline RL agent, based on a VAE model trained on the offline dataset. 
(b) Fine-tuning performance on \texttt{halfcheetah-medium} task when using online samples exclusively (\textbf{Online only}), or when using both offline and online data drawn uniformly at random (\textbf{Uniform}).
(c) Fine-tuning performance on \texttt{halfcheetah-random} and \texttt{halfcheetah-medium-expert} tasks, when using a pessimistic (denoted \textbf{CQL-init}) and a non-pessimistic (denoted \textbf{FQE-init}) Q-function, respectively. 
}
\label{fig:observation}
\vspace{-0.1in}
\end{figure}

\noindent {\bf Conservative Q-learning.}
Offline RL algorithms are off-policy RL algorithms that utilize static datasets for training an agent. In particular, conservative Q-learning [CQL; \citealp{kumar2020conservative}] pessimistically evaluates the current policy, and learns a lower bound (in expectation) of the ground-truth Q-function. To be specific, policy evaluation step of CQL minimizes the following:
\begin{align}
\label{eq:cql}
\mathcal{L}^{\tt{CQL}}_{\tt{critic}}(\theta) = \frac1{2} \underset{(s,a,s')\sim \mathcal{B}}{\mathbb{E}}\big[(Q_{\theta}-\mathcal{B}^{\pi_{\phi}}Q_{\bar{\theta}})^{2}\big] + \alpha_{0} \underset{s\sim \mathcal{B}}{\mathbb{E}}\big[ \log \sum_{a}\exp Q(s,a) - \underset{a\sim \hat{\pi}_{\beta}}{\mathbb{E}}[Q(s,a)]\big],
\end{align}
where $\hat{\pi}_{\beta}(a_{0}|s_{0}) := \frac{\sum_{s,a\in \mathcal{B}} \mathbbm{1}[s=s_{0},a=a_{0}]}{\sum_{s\in \mathcal{B}} \mathbbm{1}[s=s_{0}]}$ is the empirical behavior policy, $\alpha_{0}$ the trade-off factor, and $\mathcal{B}^{\pi}$ the bellman operator. The first term is the usual Bellman backup, and the second term is the regularization term that decreases the Q-values for unseen actions, while increasing the Q-values for seen actions. We argue that thusly trained pessimistic Q-function is beneficial for fine-tuning as well (see Figure~\ref{fig:obs_q_init}). Policy improvement step is the same as SAC defined in (\ref{eq:sac_actor}).

\section{Fine-tuning Offline RL Agent}
\label{sec:finetune}
In this section, we investigate the distribution shift problem in offline-to-online RL. 
We first explain why an agent being fine-tuned can be susceptible to distribution shift, and why distribution shift is problematic. 
Then, we demonstrate two important design choices that decide the effect of distribution shift on fine-tuning: {\em sample selection} and {\em choice of offline Q-function}.

\subsection{Distribution Shift in Offline-to-Online RL}
\label{subsec:distshift}
In offline-to-online RL, there exists a distribution shift between $d^{\tt{on}}(s,a)$ and $d^{\tt{off}}(s,a)$, where the former denotes the state-action distribution of online samples in the online buffer $\mathcal{B}^{\tt{on}}$, and the latter that of offline samples in the offline buffer $\mathcal{B}^{\tt{off}}$. Figure~\ref{fig:dist_shift} visualizes such distribution shift. Specifically, we trained a  variational autoencoder \citep{kingma2013auto} to reconstruct state-action pairs in the \texttt{halfcheetah-random} dataset that contains uniform random policy rollouts in the halfcheetah environment. Then, we compared the log-likelihood of (a) offline samples and (b) online samples collected by a CQL agent trained on the same dataset. There is a clear difference between offline and online state-action distributions. 

Such distribution shift is problematic, for the agent will enter the unseen state-action regime, where Q-values (hence value estimates used for bootstrapping) can be very inaccurate. 
Updates in such unseen regime results in erroneous policy evaluation and arbitrary policy updates, which destroys the good initial policy obtained via offline RL. 
Distribution shift can be especially severe in offline-to-online RL, for the offline RL agent is often much more performant than the behavior policy (e.g., CQL can train a medium-level agent capable of running, using transitions generated by a random policy only). 
Also, when the offline dataset is narrowly distributed, e.g., when it is generated by a single policy, the agent is more prone to distribution shift, for the agent easily deviates from the narrow, seen distribution. 

\subsection{Sample Selection}
\label{subsec:sampleselection}
In light of the above discussion, we study how sample selection affects fine-tuning.
We find that online samples, which are essential for fine-tuning, are also potentially dangerous OOD samples due to distribution shift.
Meanwhile, offline samples are in-distribution and safe, but leads to slow fine-tuning. 
As a concept experiment, we trained an agent offline via CQL (\ref{eq:cql}, \ref{eq:sac_actor}) on the \texttt{halfcheetah-medium} dataset containing medium-level transitions, 
then fine-tuned the agent via SAC (\ref{eq:sac_critic}, \ref{eq:sac_actor}).
We see that using online samples exclusively for updates (denoted \textbf{Online only} in Figure~\ref{fig:obs_sample}) leads to unstable fine-tuning, where the average return drops from about 4500 to below 3000. 
This demonstrates the harmful effect of distribution shift, where novel, OOD samples collected online cause severe bootstrap error. 

On the other hand, when using a single replay buffer for both offline and online samples then sampling uniformly at random (denoted \textbf{Uniform} in Figure~\ref{fig:obs_sample}), the agent does not use enough online samples for updates, especially when the offline dataset is large. 
As a result, value propagation is slow, and as seen in Figure~\ref{fig:obs_sample}, this scheme achieves initial stability at the cost of asymptotic performance. 
This motivates a balanced replay scheme that modulates the trade-off between using online samples (useful, but potentially dangerous), and offline samples (stable, but slow fine-tuning).

\subsection{Choice of Offline Q-function}
\label{subsec:qfunction}
Another important design choice in offline-to-online RL is the offline training of Q-function. 
In particular, we show that a pessimistically trained Q-function mitigates the effect of distribution shift, by staying conservative about OOD actions in the initial training phase. 
As a concept experiment, we compared the fine-tuning performance when using a pessimistically trained Q-function and when using a Q-function trained without any pessimistic regularization. 
Specifically, for a given offline dataset, we first trained a policy $\pi_{\phi}$ and its pessimistic Q-function $Q_{\theta_{\tt{CQL}}}$ via CQL (\ref{eq:cql}). 
Then we trained a non-pessimistic Q-function $Q_{\theta_{\tt{FQE}}}$ of the pre-trained offline policy $\pi_{\phi}$ via Fitted Q Evaluation [FQE; \citealp{paine2020hyperparameter}], an off-policy policy evaluation method that trains a given policy's Q-function. 
Finally, we fine-tuned $\{\pi_{\phi}, Q_{\theta_{\tt{CQL}}}\}$ and $\{\pi_{\phi}, Q_{\theta_{\tt{FQE}}}\}$ via SAC (\ref{eq:sac_critic}, \ref{eq:sac_actor}). See Section~\ref{supp:concept} for more details.

As shown in Figure~\ref{fig:obs_q_init}, both pessimistic and non-pessimistic Q-functions show similar fine-tuning performance on the \texttt{random} dataset, which contains random policy rollouts with good action space coverage. 
However, when fine-tuning an offline RL agent trained on the \texttt{medium-expert} dataset, which contains transitions obtained by a mixture of more selective and performant policies, non-pessimistic Q-function loses the good initial policy, reaching zero average return at one point. 

The reason is that $Q_{\tt{FQE}}$ can be overly optimistic about OOD actions at novel states when bootstrapping from them. In turn, the policy may prefer potentially bad actions, straying further away from the safe, seen trajectory.
On the other hand, $Q_{\tt{CQL}}$ remains pessimistic in the states encountered online initially, for (1) these states are incrementally different from seen states, and (2) Q-function will thus have similar pessimistic estimates due to generalization.
This points to a strategy where we first train a pessimistic Q-function offline, then let it gradually lose the pessimism as the agent gains access to a balanced mix of offline and online samples via balanced replay during fine-tuning. 
Furthermore, since a single agent's Q-function may not be pessimistic enough, we may train multiple agents offline in parallel, then deploy online the ensemble agent equipped with a higher-resolution pessimism of the Q-ensemble (see Section~\ref{subsec:qensemble} for a more detailed explanation).


\begin{wrapfigure}{1}{0.575\textwidth}
\vspace{-0.525in}
\centering
\includegraphics[width=0.575\textwidth]{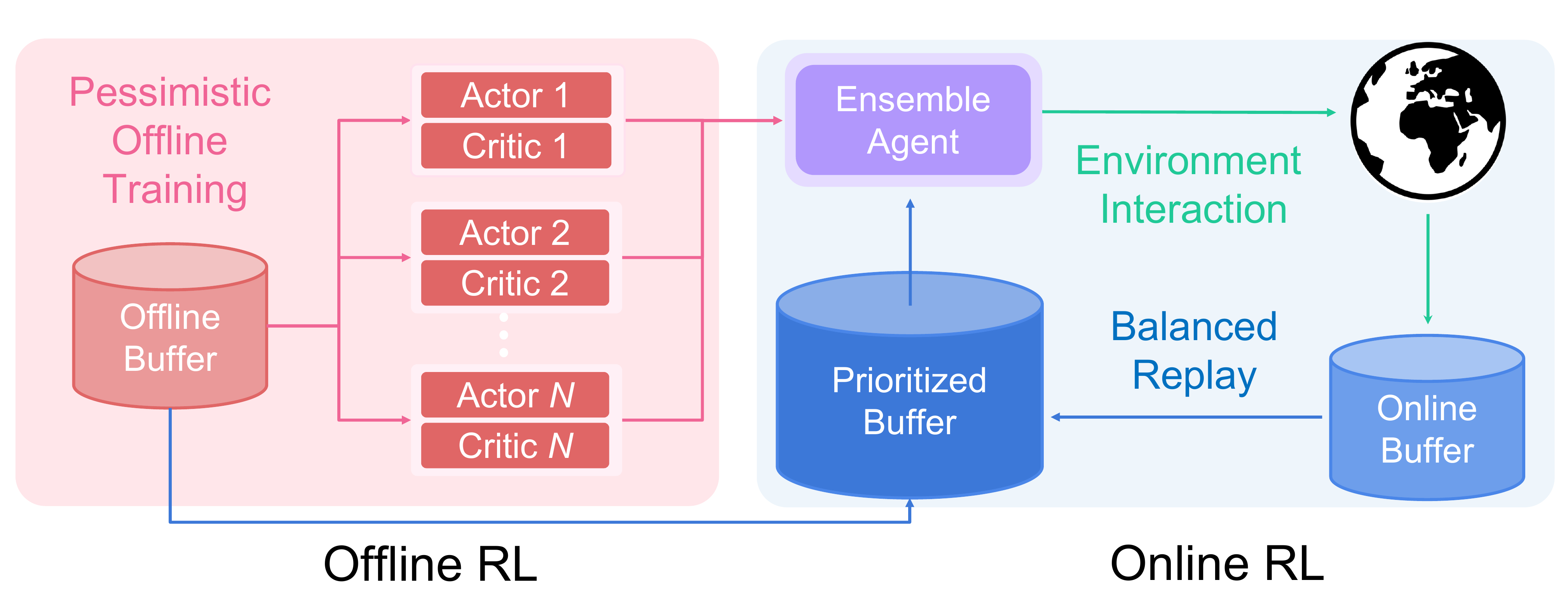}
\vspace{-0.15in}
\caption{Illustration of our framework. We first train an ensemble of $N$ CQL agents on the offline dataset. Then we fine-tune the ensemble agent using both offline and online transitions via balanced replay. In particular, we train a density ratio estimator that measures the online-ness of a given sample, then store all samples in the prioritized replay buffer with their respective density ratios as priority values. In turn, samples are drawn with probability proportional to their respective priority values. 
}
\label{fig:method}
\vspace{-0.25in}
\end{wrapfigure}

\vspace{-0.1in}
\section{Method}
\label{sec:method}
\vspace{-0.1in}
We propose a simple yet effective framework that addresses the state-action distribution shift described in Section~\ref{sec:finetune}. 
Our method comprises of two parts: (a) a balanced experience replay scheme, and (b) a pessimistic Q-ensemble scheme. 

\vspace{-0.1in}
\subsection{Balanced Experience Replay}
\label{subsec:replay}
\vspace{-0.05in}
We introduce a balanced replay scheme that enables us to safely utilize online samples by leveraging relevant, near-on-policy offline samples. 
By doing so, we can widen the sampling distribution for updates around the on-policy samples and enable timely value propagation.
The challenge here is how to design a scheme that locates and retrieves such relevant, near-on-policy samples from the offline dataset, which can often be huge.  
To achieve this, we measure the \textit{online-ness} of all available samples, and prioritize the samples according to this measure. 

In particular, when updating the agent, we propose to sample a transition $(s,a,s^\prime)\in \mathcal{B}^{\tt{off}} \cup  \mathcal{B}^{\tt{on}}$ with a probability proportional to the density ratio $w(s,a):= d^{\tt{on}}(s,a) / d^{\tt{off}}(s,a)$ of the given sample. 
This way, we can retrieve a relevant, near-on-policy sample $(s,a,s^\prime) \in \mathcal{B}^{\tt{off}}$ by locating a transition with high density ratio $w(s,a)$.
However, estimating the likelihoods $d^{\tt{off}}(s,a)$ and $d^{\tt{on}}(s,a)$ is difficult, since they can in principle be stationary distributions of complex policy mixtures\footnotemark.
To avoid this problem, we utilize a likelihood-free density ratio estimation method that estimates $w(s,a)$ by training a network $w_{\psi}(s,a)$ parametrized by $\psi$, solely based on samples from $\mathcal{B}^{\tt{off}}$ and $\mathcal{B}^{\tt{on}}$.

\footnotetext{We remark that $d^{\tt{off}}(s,a)$ is the stationary distribution of the (arbitrary) behavior policy that generated $\mathcal{B}^{\tt{off}}$, and $d^{\tt{on}}(s,a)$ the stationary distribution of the policy that generated $\mathcal{B}^{\tt{on}}$, which corresponds to the mixture of online policies observed over the course of fine-tuning.}

\noindent {\bf Training details.}
Here we describe the training procedure for the density ratio estimator $w_{\psi}(s,a)$ in detail. 
Following the idea of \citet{sinha2021experience}, we use the variational representation of f-divergences \citep{nguyen2008estimating}. 
Let $P$ and $Q$ be probability measures defined on some measurable space $\mathcal{X}$, with $P$ absolutely continuous w.r.t. $Q$, and $f(y) := y\log \frac{2y}{y+1} + \log \frac{2}{y+1}$. 
Then the Jensen-Shannon (JS) divergence is defined as $D_{JS}(P||Q) = \int_{\mathcal{X}} f(dP(x)/dQ(x))dQ(x)$. 
We then estimate the density ratio $dP/dQ$ with a parametric model $w_{\psi}(x)$, by maximizing the lower bound of $D_{JS}(P||Q)$ \citep{nguyen2008estimating}:
\begin{align}
\label{eq:weight_net}
\mathcal{L}^{\tt{DR}}(\psi) &= \mathbb{E}_{x \sim P}[f'(w_{\psi}(x))] - \mathbb{E}_{x\sim Q}[f^{*}(f'(w_{\psi}(x)))],
\end{align}
where $w_{\psi}(x) \geq 0$ is parametrized by a neural network whose outputs are forced to be non-negative via activation functions, and $f^{*}$ denotes convex conjugate. In particular, we obtain an estimate $w_{\psi}(s,a)$ of $d^{\texttt{on}}/d^{\texttt{off}}$ by considering probability distributions $P$ and $Q$ with densities $d^{\texttt{on}}$ and $d^{\texttt{off}}$, respectively. 
In practice, we sample from $\mathcal{B}^{\tt{on}}$ for the first term in (\ref{eq:weight_net}), and from $\mathcal{B}^{\tt{off}}$ for the latter.
For more stable density ratio estimates, we apply self-normalization \citep{cochran2007} to the estimated density ratios over $\mathcal{B}^{\tt{off}}$,  similar to \citet{sinha2021experience}. More details can be found in Section~\ref{supp:setup}.

\vspace{-0.1in}
\subsection{Pessimistic Q-Ensemble}
\label{subsec:qensemble}
\vspace{-0.1in}
In order to mitigate distribution shift more effectively, we leverage multiple pessimistically trained Q-functions. We consider an ensemble of $N$ CQL agents pre-trained via update rules (\ref{eq:sac_actor}, \ref{eq:cql}), i.e., $\{Q_{\theta_{i}}, \pi_{\phi_{i}}\}_{i=1}^{N}$, where $\theta_{i}$ and $\phi_{i}$ denote the parameters of the $i$-th agent's Q-function and policy, respectively. Then we use the ensemble of actor-critic agents whose Q-function and policy are defined as follows:
\begin{align}
\label{eq:q_ensemble}
Q_{\theta} := \frac{1}{N} \sum_{i=1}^{N} Q_{\theta_{i}}, 
\quad \;
\pi_{\phi}(\cdot|s) = \mathcal{N}\bigg(\frac1{N}\sum_{i=1}^{N} \mu_{\phi_{i}}(s), \quad
\frac{1}{N} \sum_{i=1}^{N} (\sigma^{2}_{\phi_{i}}(s) + \mu_{\phi_{i}}^{2}(s)) - \mu_{\phi}^{2}(s)\bigg),
\end{align}
where $\theta := \{\theta_{i}\}_{i=1}^{N}$ and $\phi := \{\phi_{i}\}_{i=1}^{N}$. 
Note that the policy is simply modeled as Gaussian with mean and variance of the Gaussian mixture policy $\frac1{N}\sum_{i=1}^{N}\pi_{\phi_{i}}$.
In turn, $\theta$ and $\phi$ are updated via update rules (\ref{eq:sac_critic}) and (\ref{eq:sac_actor}), respectively, during fine-tuning. 

By using a pessimistic Q-function, the agent remains pessimistic with regard to the unseen actions at states encountered online during initial fine-tuning. This is because during early fine-tuning, states resemble those present in the offline dataset, and Q-function generalizes to these states. 
As we show in our experiments, this protects the good initial policy from severe bootstrap error. 
And by leveraging multiple pessimistically trained Q-functions, we obtain a more high-resolution pessimism about the unseen data regime. 
That is, when an individual Q-function may erroneously have high values for unseen samples, Q-ensemble is more robust to these individual errors, and more reliably assigns lower values to unseen samples.
Computational overhead of ensemble is discussed in Section~\ref{supp:complexity}.

\vspace{-0.1in}
\section{Related work}
\vspace{-0.1in}
\label{sec:related}
\noindent {\bf Offline RL.} Offline RL algorithms aim to train RL agents exclusively with pre-collected datasets.
To address the state-conditional action distribution shift, prior methods (a) explicitly constrain the policy to be closed to the behavior policy \citep{fujimoto2019off, kumar2019stabilizing, wu2019behavior, siegel2020keep, ghasemipour2020emaq}, or (b) train pessimistic value functions \citep{kidambi2020morel, yu2020mopo, kumar2020conservative}. In particular, CQL \citep{kumar2020conservative} was used to learn various robotic manipulation tasks \citep{singh2020cog}. We also build on CQL, so as to leverage pessimism regarding data encountered online during fine-tuning. 

\vspace{-0.05in}
\noindent {\bf Online RL with offline datasets.}
Several works have explored employing offline datasets for online RL to improve sample efficiency.
Some assume access to demonstration data \citep{ijspeert2003learning, kim2013learning, rajeswaran2017learning, zhu2019dexterous,vecerik2017leveraging}, 
which is limited in that they assume optimality of the dataset. 
To overcome this, \citet{nair2021awac} proposed Advantage Weighted Actor Critic (AWAC), which performs regularized policy updates so that the policy stays close to the observed data during both offline and online phases. 
We instead advocate adopting pessimistic \textit{initialization}, such that we may prevent overoptimism and bootstrap error in the initial online phase, and lift such pessimism once unnecessary, as more online samples are gathered.
Some recent works extract behavior primitives from offline data, then learn to compose them online \citep{pertsch2020spirl,ajay2021opal,singh2021parrot}. It would be interesting to apply our method in these setups. 

\vspace{-0.05in}
\noindent {\bf Experience replay.}
The idea of retrieving important samples for RL was introduced in \citet{schaul2015prioritized}, where they prioritize samples with high temporal-difference error.
The work closest to ours is \citet{sinha2021experience}, which utilizes the density ratios between off-policy and near-on-policy state-action distributions as importance weights for policy evaluation.
Our approach differs in that we utilize density ratios for retrieving relevant samples from the offline dataset. 

\vspace{-0.05in}
\noindent {\bf Ensemble methods.} 
In the context of model-free RL, ensemble methods have been studied for addressing Q-function's overestimation bias \citep{hasselt2010double, hasselt2016deep, anschel2017averaged, fujimoto2018addressing, lan2020maxmin}, for better exploration \citep{osband2016deep,chen2017ucb,lee2021sunrise}, or for reducing bootstrap error propagation \citep{lee2020sunrise}. 
The closest to our approach is \citet{anschel2017averaged} that stabilizes Q-learning by using the average of previously learned Q-values as the target Q-value. 
While prior works mostly focus on online RL and estimate the ground-truth Q-functions, we leverage an ensemble of pessimistically pre-trained Q-functions for safe offline-to-online RL.


\section{Experiments}
\begin{figure*}[t!]
\vspace{-0.35in}
\centering
\hfill
\subfloat[Performance \label{fig:manip_perf}]{%
   \includegraphics[width=0.74\textwidth]{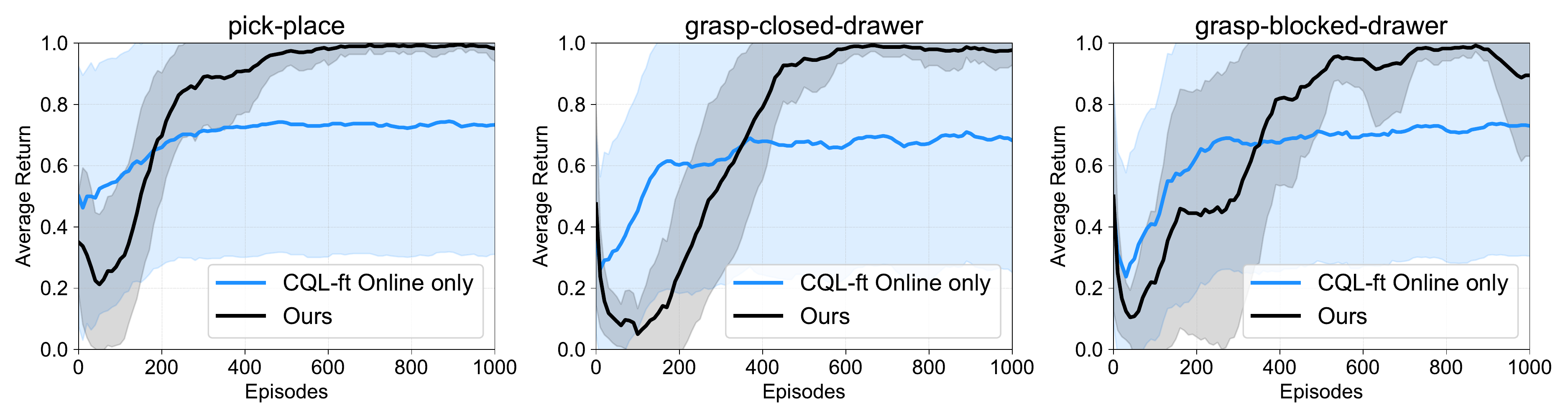}
}
\hfill
\subfloat[Buffer analysis \label{fig:manip_buffer}]{%
   \includegraphics[width=0.24\textwidth]{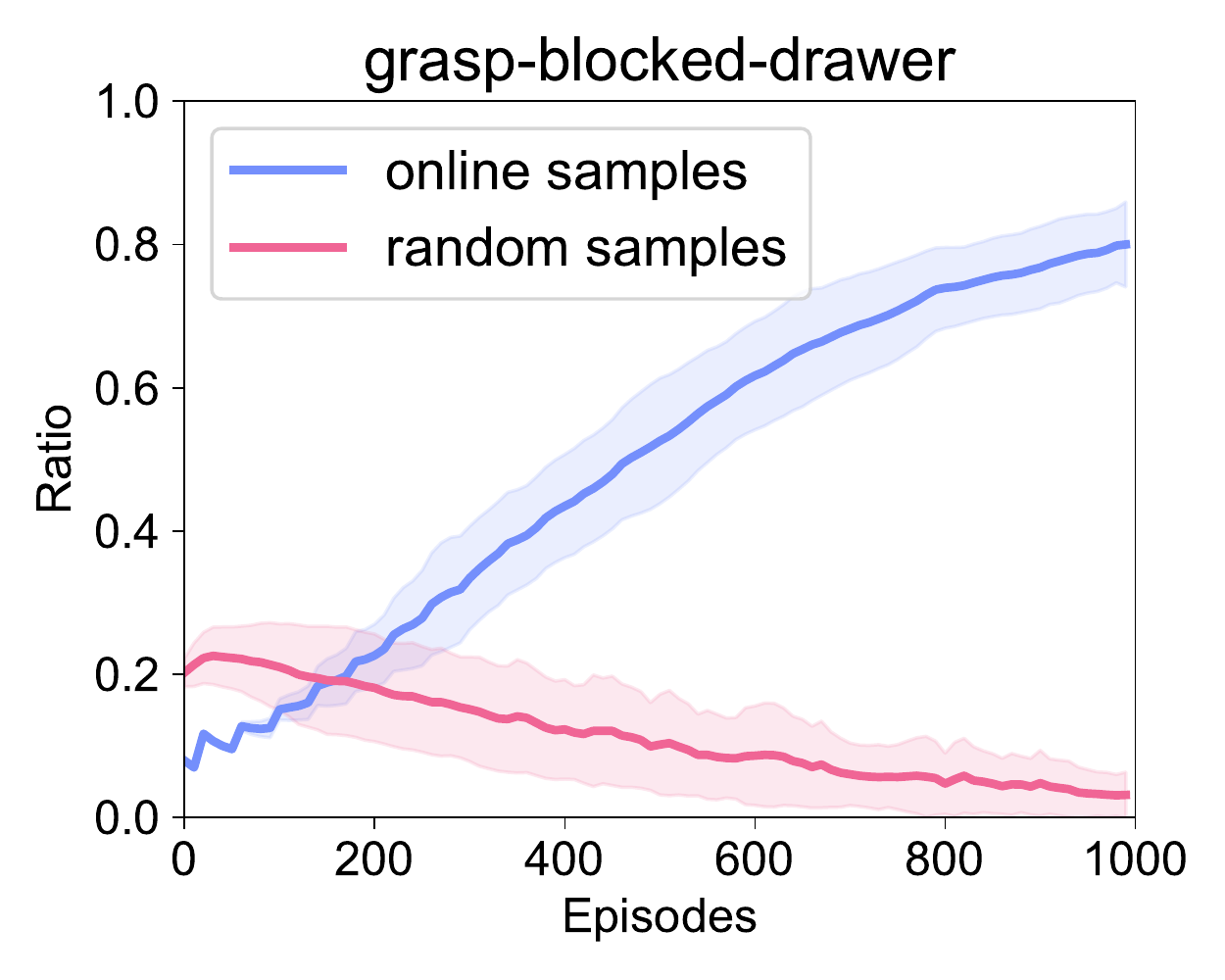}
}
\hfill
\caption{
(a) Fine-tuning performance for robotic manipulation tasks considered. 
(b) Proportion of random data used for during fine-tuning decreases over time.  
The solid lines and shaded regions represent mean and standard deviation, respectively, across eight runs.
}
\vspace{-0.15in}
\label{fig:manipulation_experiment}
\end{figure*}

\vspace{-0.1in}
\label{section:experiment}
We designed our experiments to answer the following questions:
\begin{itemize}[leftmargin=5.5mm, topsep=-2.5pt, itemsep=-2.5pt]
    \item [$\bullet$] How does our method compare to existing offline-to-online RL methods and an online RL method that learns from scratch (see Figure~\ref{fig:performance})?
    \item [$\bullet$] Can our balanced replay scheme locate offline samples relevant to the current policy (see Figure~\ref{fig:analysis_buffer}) and improve the fine-tuning performance by utilizing these samples (see Figure~\ref{fig:abl_experience_replay})?
    \item [$\bullet$] Can our pessimistic Q-ensemble scheme discriminate unseen actions (see Figure~\ref{fig:abl_discriminative}) and successfully stabilize the fine-tuning procedure by mitigating distribution shift (see Figure~\ref{fig:abl_ensemble})?
    \item [$\bullet$] Does our method scale to vision-based robotic manipulation tasks (see Figure~\ref{fig:manipulation_experiment})?
\end{itemize}

\subsection{Locomotion Tasks}
\label{subsec:locomotion}
\noindent {\bf Setup.} We consider MuJoCo \citep{todorov2012mujoco} locomotion tasks, i.e., \texttt{halfcheetah}, \texttt{hopper}, and \texttt{walker2d}, from the D4RL benchmark suite \citep{fu2020d4rl}.
To demonstrate the applicability of our method on various suboptimal datasets, we use four dataset types: \texttt{random}, \texttt{medium}, \texttt{medium-replay}, and \texttt{medium-expert}. Specifically, \texttt{random} and \texttt{medium} datasets contain samples collected by a random policy and a medium-level policy, respectively.
\texttt{medium-replay} datasets contain all samples encountered while training a medium-level agent from scratch, and \texttt{medium-expert} datasets contain samples collected by both medium-level and expert-level policies.
For our method, we use ensemble size $N=5$. More experimental details are provided in Section~\ref{supp:setup}.

\begin{figure*}[t!]
\vspace{-0.25in}
\centering
\includegraphics[width=0.65\textwidth]{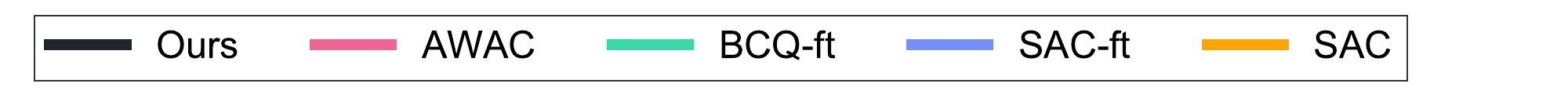}
\\
\includegraphics[width=0.95\textwidth]{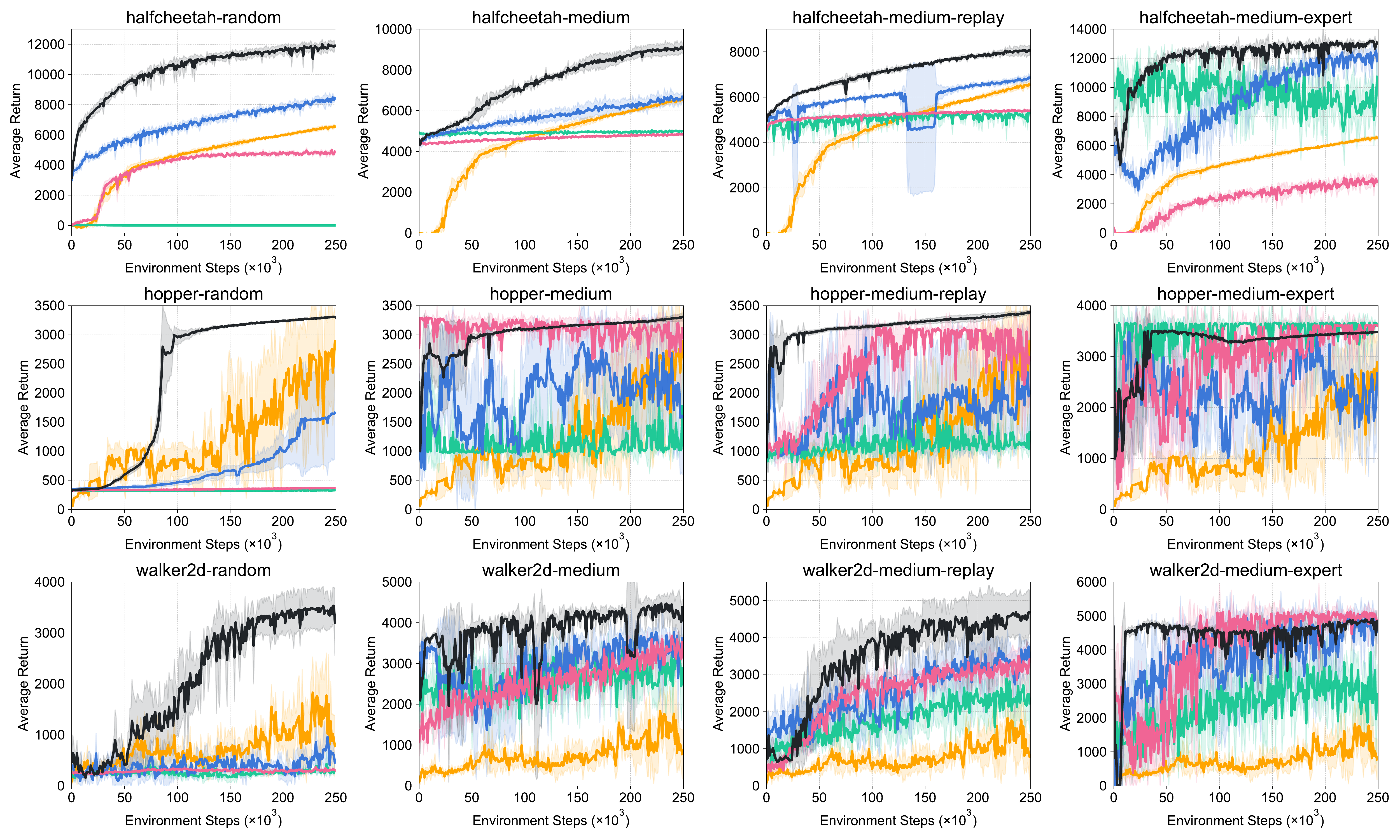}
\caption{ Performance on D4RL \citep{fu2020d4rl} MuJoCo locomotion tasks during online fine-tuning. The solid lines and shaded regions represent mean and standard deviation, respectively, across four runs.
}
\vspace{-0.15in}
\label{fig:performance}
\end{figure*}

\begin{figure*}[t!]
\vspace{-0.35in}
\centering
\hfill
\subfloat[Buffer analysis]{%
   \includegraphics[width=0.24\textwidth]{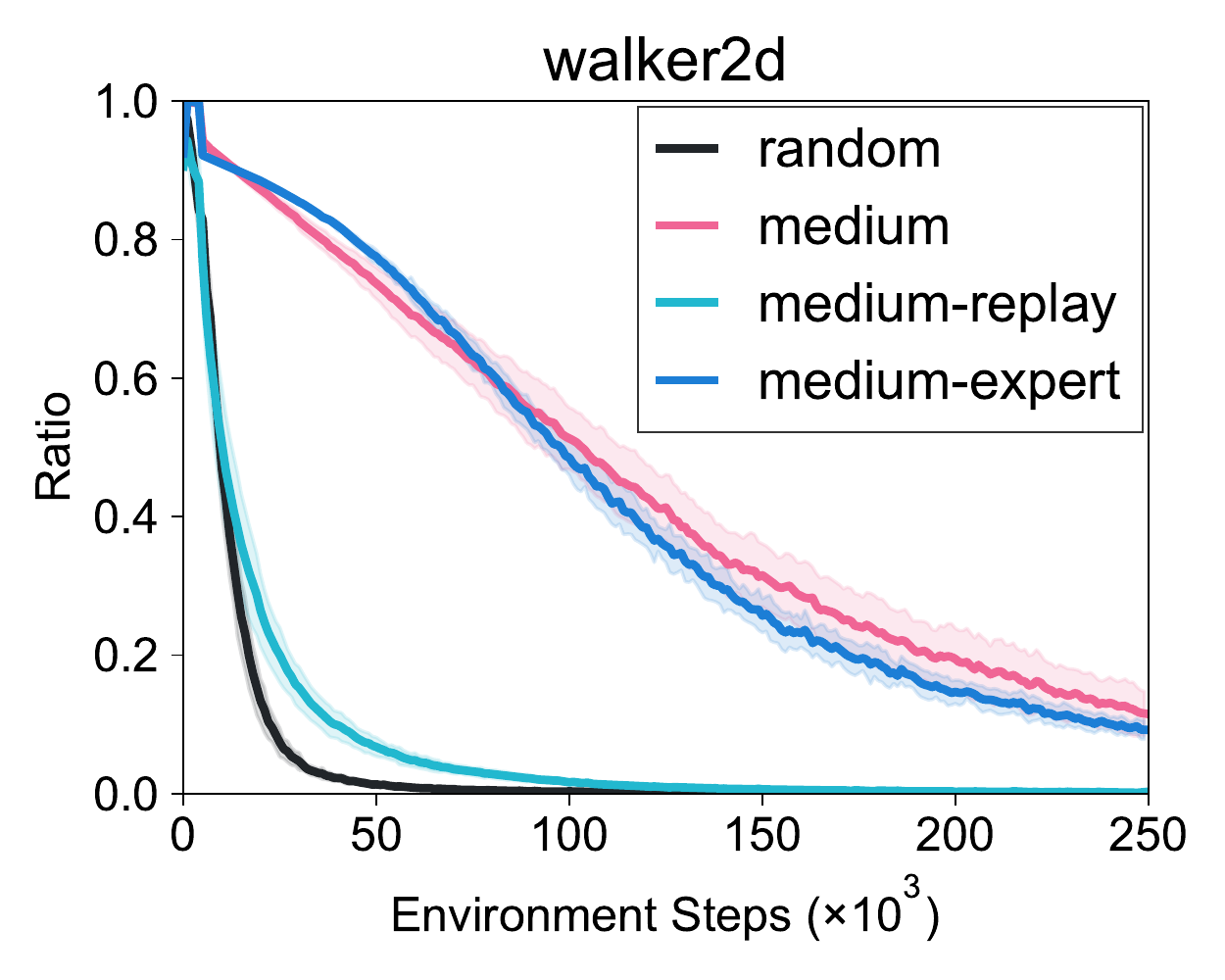}
   \label{fig:analysis_buffer}
}
\hfill
\subfloat[Ensemble analysis]{%
   \includegraphics[width=0.24\textwidth]{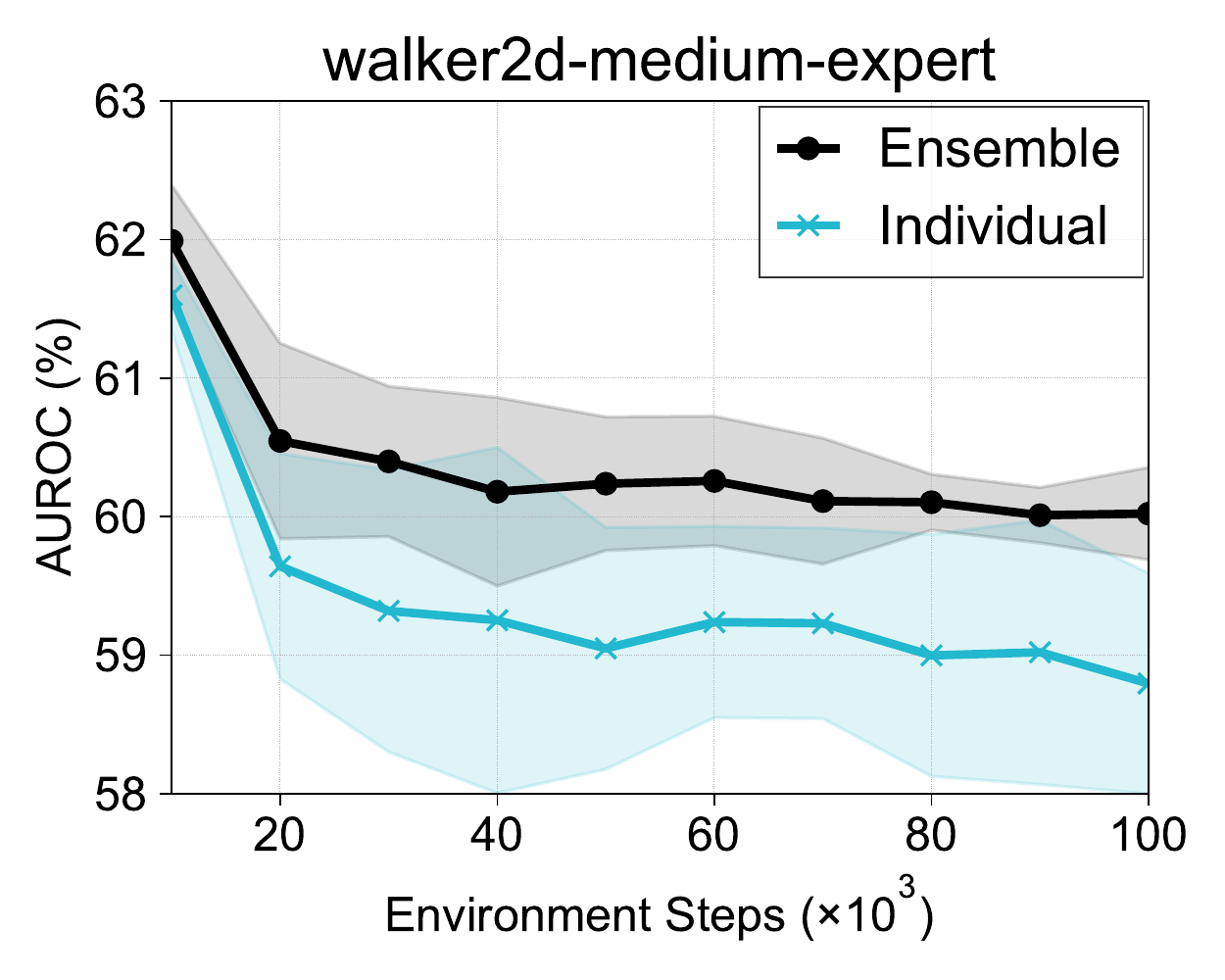}
   \label{fig:abl_discriminative}
}
\hfill
\subfloat[Buffer ablation]{%
   \includegraphics[width=0.24\textwidth]{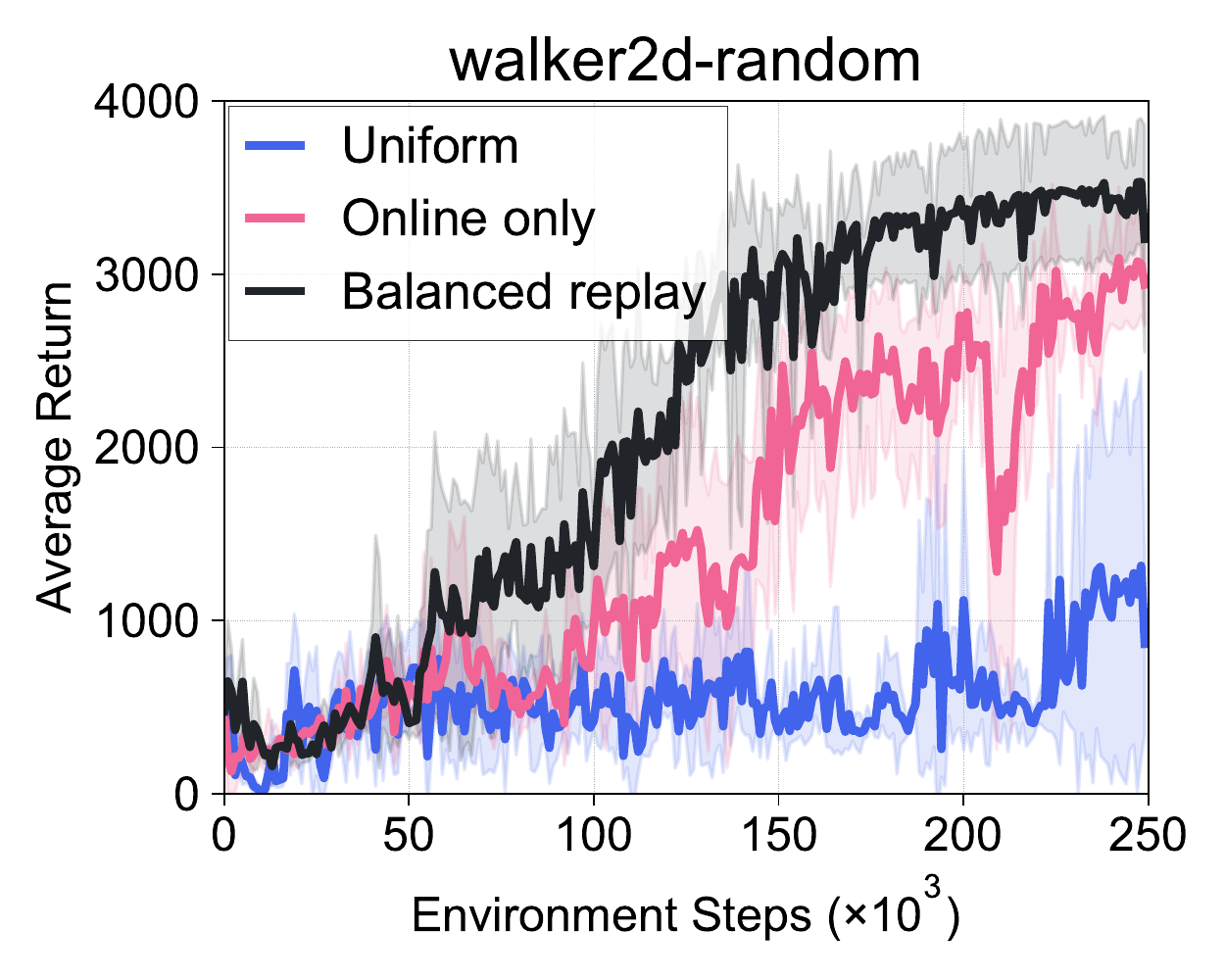}
   \label{fig:abl_experience_replay}
}
\hfill
\subfloat[Ensemble ablation]{%
   \includegraphics[width=0.24\textwidth]{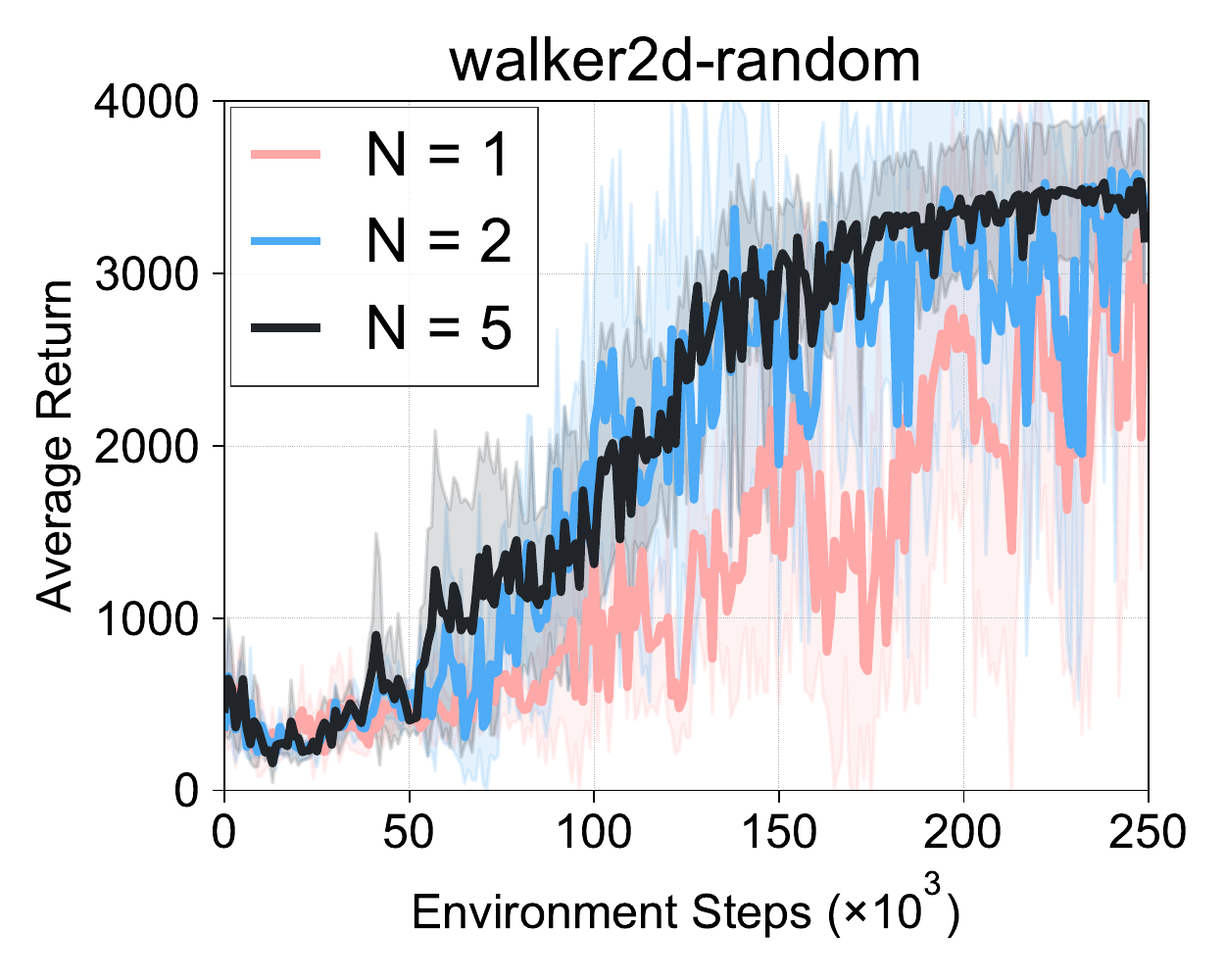}
   \label{fig:abl_ensemble}
}
\hfill
\vspace{-0.0in}
\caption{
(a) Proportion of offline samples used for updates as the agent is fine-tuned online, for \texttt{walker2d} tasks. 
(b) AUROC (\%) over the course of fine-tuning on \texttt{walker2d-medium-expert}, where the Q-function is interpreted as a binary classifier that classifies a given state-action pair $(s,a)$ as either a seen pair $(s, a_{\tt{seen}})$ or an unseen pair $(s,a_{\tt{uniform}})$, for a state $s$ encountered online.
Pessimistic Q-ensemble shows a stronger discriminative ability.
(c) Performance on \texttt{walker2d-random} with and without balanced experience replay. We consider two setups where balanced experience replay is not used: (i) Uniform, where offline and online samples are sampled uniformly from the same buffer for updates, and (ii) Online only, where the offline agent is fine-tuned using online samples only.
(d) Performance on \texttt{walker2d-random} with varying ensemble size $N\in \{1,2,5\}$. 
The solid lines and shaded regions represent mean and standard deviation, respectively, across four runs.}
\label{fig:ablation_and_analysis}
\vspace{-0.175in}
\end{figure*}

\noindent {\bf Comparative Evaluation.}
\label{subsec:exp_comparative}
We consider the methods outlined below as baselines for comparative evaluation. For fair comparison, we applied ensemble to all baselines except SAC-ft, since the results for SAC-ft with ensemble can be found in the ablation studies (see Figure~\ref{fig:abl_experience_replay}).
\begin{itemize}[leftmargin=5.5mm, topsep=-2.5pt, itemsep=-2.5pt]
    \item Advantage Weighted Actor Critic [AWAC; \citealp{nair2021awac}]: 
     an offline-to-online RL method that trains the policy to imitate actions with high advantage estimates. 
    \item BCQ-ft: Batch-Constrained deep Q-learning [BCQ; \citealp{fujimoto2019off}], is an offline RL method that updates policy by modeling the behavior policy using a conditional VAE \citep{sohn2015learning}. We extend BCQ to the online fine-tuning setup by applying the same update rules as offline training. 
    \item SAC-ft: Starting from a CQL agent trained via (\ref{eq:cql}, \ref{eq:sac_actor}), we fine-tune the agent via SAC updates (\ref{eq:sac_critic}, \ref{eq:sac_actor}). Justification for excluding the CQL regularization term from (\ref{eq:cql}) during fine-tuning can be found in Section~\ref{supp:additional_exp_locomotion}.
    \item SAC: a SAC agent trained from scratch via (\ref{eq:sac_critic}, \ref{eq:sac_actor}), i.e., the agent has no access to the offline dataset. This baseline highlights the benefit of offline-to-online RL, as opposed to fully online RL, in terms of sample efficiency. 
\end{itemize}

Figure~\ref{fig:performance} shows the performances of our method and baseline methods considered during the online RL phase. 
In most tasks, our method outperforms the baseline methods in terms of both sample-efficiency and final performance.
In particular, our method significantly outperforms SAC-ft, which shows that balanced replay and pessimistic Q-ensemble are indeed essential. 

We also emphasize that our method performs consistently well across all tasks, while the performances of AWAC and BCQ-ft are highly dependent on the quality of the offline dataset.
For example, we observe that AWAC and BCQ-ft show competitive performances in tasks where the datasets are generated by high-quality policies, i.e., \texttt{medium-expert} tasks, but perform worse than SAC on \texttt{random} tasks. 
This is because AWAC and BCQ-ft employ the same regularized, pessimistic update rule for offline and online setups alike, either explicitly (BCQ-ft) or implicitly (AWAC), which leads to slow fine-tuning. 
Our method instead relies on pessimistic \textit{initialization}, and hence enjoys much faster fine-tuning, while not sacrificing the initial training stability.

\noindent {\bf Balanced replay analysis.}
To investigate the effectiveness of our balanced experience replay scheme for locating near-on-policy samples in the offline dataset, we report the ratios of offline samples used for updates fine-tuning proceeds.
Figure~\ref{fig:analysis_buffer} shows that for the \texttt{random} task, offline samples quickly become obsolete, as they quickly become irrelevant to the policy being fine-tuned. However, for the \texttt{medium-expert} task, offline samples include useful expert-level transitions that are relevant to the current policy, hence are replayed throughout the online training. This shows that our balanced replay scheme is capable of utilizing offline samples only when appropriate. 

\noindent {\bf Q-ensemble analysis.}
\label{subsubsec:q_ensemble_analysis}
We quantitatively demonstrate that pessimistic Q-ensemble indeed provides more discriminative value estimates, i.e., having distinguishably lower Q-values for unseen actions than for seen actions. 
In particular, we consider a \texttt{medium-expert} dataset, where the offline data distribution is narrow, and the near-optimal offline policy can be brittle. 
Let $\mathcal{D}^{\tt{real}}_{T} := \{(s_{i},a_{i})\}_{i=1}^{T}$ be the samples collected online up until timestep $T$. 
We construct a ``fake" dataset by replacing the actions in $\mathcal{D}^{\tt{real}}_{T}$ with random actions, i.e., $\mathcal{D}^{\tt{fake}}_{T} := \{(s_{i},a_{\tt{unif}})\}_{i=1}^{T}$. 
Interpreting $Q(s,a)$ as the confidence value for classifying real and fake transitions, we measure the area under ROC (AUROC) curve values over the course of fine-tuning. 
As seen in Figure~\ref{fig:abl_discriminative}, Q-ensemble demonstrates superior discriminative ability, which leads to stable fine-tuning.

\clearpage

\noindent {\bf Ablation studies.}
Figure~\ref{fig:abl_experience_replay} shows that balanced replay improves fine-tuning performance by sampling near-on-policy transitions. On the other hand, two other na\"ive sampling schemes -- (a) Uniform, where offline and online samples are sampled uniformly from the same buffer, and (b) Online only, where the offline agent is fined-tuned using online samples exclusively -- suffer from slow and unstable improvement, even with pessimistic Q-ensemble.
This shows that balanced replay is crucial for reducing the harmful effects of distribution shift. 

Also, Figure~\ref{fig:abl_ensemble} shows that fine-tuning performance improves as the ensemble size $N$ increases, which shows that larger ensemble size provides higher-resolution pessimism, leading to more stable policy updates. Ablation studies for all tasks can be found in Section~\ref{supp:additional_exp_locomotion}.

\subsection{Robotic Manipulation Tasks}
\label{subsec:manip}
\noindent {\bf Setup.} 
We consider three sparse-reward pixel-based manipulation tasks from \citet{singh2020cog}:
(1) \texttt{pick-place}: pick an object and put it in the tray; 
(2) \texttt{grasp-closed-drawer}: grasp an object in the initially closed bottom drawer;  
(3) \texttt{grasp-blocked-drawer}: grasp an object in the initially closed bottom drawer, where the initially open top drawer blocks the handle for the bottom drawer. 
Episode lengths for the tasks are 40, 50, 80, respectively. 

The original dataset \citep{singh2020cog} for each task consists of scripted exploratory policy rollouts. For example, for \texttt{pick-place}, the dataset contains scripted pick attempts and place attempts. 
However, it is rarely the case that logged data `in the wild' contains such structured, high-quality transitions only. 
We consider a more realistic setup where the dataset also includes undirected, exploratory samples --
we replace a subset of the original dataset with uniform random policy rollouts. Note that random policy rollouts are common in robotic tasks \citep{finn2017deep,ebert2018visual}. 
We used ensemble size $N=4$ for our method. 
More details about the tasks and dataset construction are provided in Section~\ref{supp:manipulation_setup}.

\noindent {\bf Comparative Evaluation.}
We compare our method with the method considered in \citet{singh2020cog}, namely, CQL fine-tuning with online samples only. 
CQL-ft fails to solve the task in some of the seeds, resulting in high variance as seen in Figure~\ref{fig:manip_perf}.
This is because CQL shows inconsistent offline performance across random seeds due to such factors as 
difficulty of training on mixture data \citep{fu2020d4rl}, 
instability of offline agents over stopping point of training \citep{fujimoto2021minimalist}, 
and sparsity of rewards. 
With no access to (pseudo-)expert offline data and due to heavy regularization of CQL, such CQL agents hardly improve. 
Meanwhile, our method consistently learns to perform the task within a reasonable amount of additional environment interaction (40K to 80K steps).

\noindent {\bf Buffer analysis.}
We analyze whether balanced replay scales to image-based robotic tasks. 
As seen in Figure~\ref{fig:manip_buffer}, without any privileged information, balanced replay automatically selects relevant offline samples for updates, while filtering out task-irrelevant, random data as fine-tuning proceeds.


\section{Conclusion}
\label{sec:conclusion}
In this paper, we identify state-action distribution shift as the major obstacle in offline-to-online RL. 
To address this, we present a simple framework that incorporates (1) a balanced experience replay scheme, and (2) a pessimistic Q-ensemble scheme.
Our experiments show that the proposed method performs well across many continuous control robotic tasks, including locomotion and manipulation tasks. 
We expect our method to enable more sample-efficient training of robotic agents by leveraging offline samples both for offline and online learning. 
We also believe our method could prove to be useful for other relevant topics such as scalable RL \citep{kalashnikov2018qtopt} and RL safety \citep{garcia2015comprehensive}. 



\clearpage
\acknowledgments{
This work was supported by Microsoft and Institute of Information \& communications Technology Planning \& Evaluation (IITP) grant funded by the Korea government(MSIT)  
(No.2019-0-00075, Artificial Intelligence Graduate School Program(KAIST)).
We would like to thank anonymous reviewers for providing helpful feedbacks and suggestions in improving our paper.
}


\bibliography{example}  

\newpage
\appendix
\onecolumn

\begin{center}{\bf {\LARGE Supplementary Material}}
\end{center}

\section{Pseudocode}
\label{supp:pseudocode}
We provide a pseudocode for the entire training procedure. For conciseness of presentation, we henceforth let $\tau = (s,a,r,s')$ denote a transition stored in the replay buffer. 
\begin{algorithm}[!ht]
\caption{Offline RL} \label{alg:offlineRL}
\begin{algorithmic}[1]
\STATE {\bf Inputs}: Ensemble size $N$, offline dataset $\mathcal{B}^{\tt{off}}$
\vspace{1mm}
\hrule
\vspace{1mm}
\STATE Initialize parameters of $N$ independent CQL agents $\{Q_{\theta_{i}}, \pi_{\phi_{i}}\}_{i=1}^{N}$
\FOR{$i=1, \cdots, N$}
\FOR{each training iteration}
\STATE Sample a random minibatch $\{\tau_{j}\}_{j=1}^{B} \sim \mathcal{B}^{\tt{off}}$
\STATE Calculate $\mathcal{L}^{\tt{CQL}}_{\tt{critic}}(\theta_{i})$ and update $\theta_{i}$
\STATE Calculate $\mathcal{L}^{\tt{SAC}}_{\tt{actor}}(\phi_{i})$ and update $\phi_{i}$
\ENDFOR              
\ENDFOR
\STATE {{\textsc{// Ensemble Agent }}}
\STATE Define Q-ensemble $Q_{\theta} := \frac1{N}  \sum_{i=1}^{N} Q_{\theta_{i}}$ 
\STATE Define $\pi_{\phi}(\cdot|s) := \mathcal{N}\bigg(\frac1{N}\sum_{i=1}^{N} \mu_{\phi_{i}}(s), \quad
\frac{1}{N} \sum_{i=1}^{N} (\sigma^{2}_{\phi_{i}}(s) + \mu_{\phi_{i}}^{2}(s)) - \mu_{\phi}^{2}(s)\bigg)$
\STATE \textbf{Return} Ensemble agent $\{Q_{\theta}, \pi_{\phi}\}$
\end{algorithmic}
\end{algorithm}
\vspace{-0.15in}
\begin{algorithm}[!h]
\caption{Online RL} \label{alg:onlineRL}
\begin{algorithmic}[1]
\STATE {\bf Inputs}: Ensemble agent $\{\pi_{\phi}, Q_{\theta}\}$, offline buffer $\mathcal{B}^{\tt{off}}$
\vspace{1mm}
\hrule
\vspace{1mm}
\STATE Initialize parameters of density ratio estimation network, $\psi$
\STATE Initialize online buffer $\mathcal{B}^{\tt{on}} \leftarrow \emptyset$, prioritized buffer $\mathcal{B}^{\tt{priority}} \leftarrow \emptyset$, and default priority value $p_{0} \leftarrow 1.0$
\FOR{$j=1, \dots , |\mathcal{B}^{\tt{off}}|$}
\STATE $\mathcal{B}^{\tt{priority}} \leftarrow \mathcal{B}^{\tt{priority}} \cup \{(\tau_{j}, p_{0})\}$
\ENDFOR
\STATE $p_{0} \leftarrow P_{0}$ (see Training details for balanced replay in Section~\ref{supp:setup} for explanation)
\FOR{each iteration}
\STATE {{\textsc{// Collect training samples}}}
\STATE Collect a transition $\tau = (s, a, r, s')$ via environment interaction with $\pi_{\phi}$
\STATE Update $\mathcal{B}^{\tt{on}} \leftarrow \mathcal{B}^{\tt{on}} \cup \{ \tau \}$, \hspace{1mm} $\mathcal{B}^{\tt{priority}} \leftarrow \mathcal{B}^{\tt{priority}} \cup \{ (\tau, p_{0}) \}$
\STATE {{\textsc{// Update density ratio estimation network}}}
\STATE Sample $ \{ \tau_{i}^{\tt{off}} \}_{i=1}^{B} \sim \mathcal{B}^{\tt{off}}$, $\{\tau_{i}^{\tt{on}} \}_{i=1}^{B} \sim \mathcal{B}^{\tt{on}}$
\STATE Calculate $\mathcal{L}^{\tt{DR}}(\psi)$ and update $\psi$
\STATE {{\textsc{// Update agent and priority values}}}
\STATE Sample a random minibatch $\{\tau_{j}\}_{j=1}^{B} \sim \mathcal{B}^{\tt{priority}}$
\STATE Calculate $\mathcal{L}^{\tt{SAC}}_{\tt{critic}}(\theta)$ and update $\theta$
\STATE Calculate $\mathcal{L}^{\tt{SAC}}_{\tt{actor}}(\phi)$ and update $\phi$
\FOR{$j=1,\dots,B$}
\STATE Update priority of $\tau_{j}$ to be $\frac{w(s_{j},a_{j})}{\mathbb{E}_{\tau^{\tt{off}} \sim \mathcal{B}^{\tt{off}}}[w(s,a)]}$ (see Training details for balanced replay in Section~\ref{supp:setup} for explanation)
\STATE $p_{0} \leftarrow \max(p_{0}, \frac{w(s_{j},a_{j})}{\mathbb{E}_{\tau^{\tt{off}}\sim \mathcal{B}^{\tt{off}}}[w(s, a)]})$
\ENDFOR
\ENDFOR
\end{algorithmic}
\end{algorithm}

\newpage

\section{Computational Complexity}
\label{supp:complexity}
Ensemble incurs additional computation overhead for training and inference. A na\"ive implementation with ensemble size $N$ leads to $N \times$ training and inference time. However, it is possible to parallelize the networks (which we did for MuJoCo locomotion tasks) for computational efficiency.

\section{Training Details for Concept Experiments}
\label{supp:concept}
{\bf VAE training.} 
For generating the plot in Figure~\ref{fig:dist_shift}, we first split the \texttt{halfcheetah-random} offline dataset consisting of random halfcheetah transitions into training and validation datasets. Then we trained a variational autoencoder on the training dataset using the Adam optimizer \citep{kingma2014adam} with learning rate $1\mathrm{e}{-3}$. We estimated the log-likelihood of (a) online samples gathered by an offline agent trained via CQL (\ref{eq:cql}, \ref{eq:sac_actor}), and (b) offline samples in the validation dataset, using the importance weight estimator \citep{burda2015importance}.

{\bf Sample selection analysis.} 
For generating the plot in Figure~\ref{fig:obs_sample}, we first trained a CQL agent $Q_{\theta}, \pi_{\phi}$ on the \texttt{halfcheetah-medium} dataset consisting of medium-quality halfcheetah transitions. Then, we fine-tune the CQL agent with two different strategies: Uniform and Online only. 
First, Uniform refers to the strategy where we keep a single replay buffer, in which we store both offline and online samples, then sampling a minibatch from the replay buffer uniformly at random, i.e., $\{\tau_{j}\}_{j=1}^{B} \sim \mathcal{B}^{\tt{off}} \cup \mathcal{B}^{\tt{on}}$. Here, the sampling procedure is agnostic of whether the sample is offline or online. 
On the other hand, Online only refers to the strategy that exploits online samples only, i.e., $\{\tau_{j}\}_{j=1}^{B} \sim \mathcal{B}^{\tt{on}}$. 

{\bf Choice of offline Q-function.} 
For generating the plots in Figure~\ref{fig:obs_q_init}, we considered the \texttt{halfcheetah-random} and \texttt{halfcheetah-medium-expert} offline datasets, each consisting of random, and a mixture of medium and expert transitions, respectively.
For a given offline dataset $\mathcal{B}^{\tt{off}}$, we first trained a CQL agent $\{Q_{\theta_{\tt{CQL}}}, \pi_{\phi}\}$ via (\ref{eq:cql}, \ref{eq:sac_actor}).  
Then, with $\pi_{\phi}$ and $\mathcal{B}^{\tt{off}}$, we trained a non-pessimistic $Q_{\theta_{\tt{FQE}}}$ via Fitted Q-Evaluation [FQE; \citealp{paine2020hyperparameter}]. To elaborate, given a fixed policy $\pi_{\phi}$, FQE trains a Q-function starting from random initialization by performing policy evaluation (Bellman backup) until convergence. In practice, we used the Adam optimizer \citep{kingma2014adam} with learning rate $3\mathrm{e}{-4}$ and batch size $B = 256$, and trained for 250k iterations. 
In the online fine-tuning phase, we fine-tuned $\{\pi_{\phi}, Q_{\theta_{\tt{CQL}}}\}$ and $\{\pi_{\phi}, Q_{\theta_{\tt{FQE}}}\}$, respectively, via SAC update rules (\ref{eq:sac_critic}, \ref{eq:sac_actor}). We applied the balanced replay scheme introduced in Section~\ref{subsec:replay}, in order to facilitate fine-tuning.

\newpage

\section{Experimental Setup Details}
\label{supp:setup}

\subsection{Locomotion Tasks}
\label{supp:locomotion_setup}

\begin{wrapfigure}{r}{0.45\textwidth}
\vspace{-0.7in}
\centering
\includegraphics[width=0.45\textwidth]{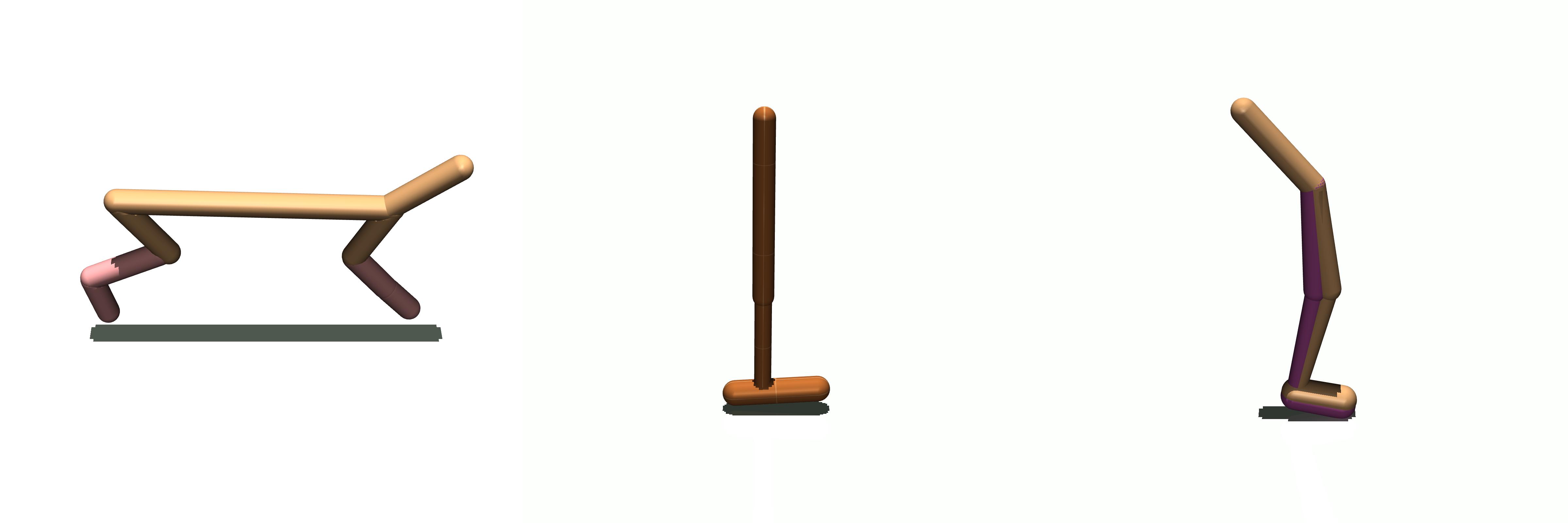}
\vspace{-0.3in}
\caption{Locomotion tasks. \texttt{halfcheetah}, \texttt{hopper}, and \texttt{walker2d}, in order.}
\end{wrapfigure}

\textbf{Task detail \& offline dataset.} 
For training offline RL agents, we used the publicly available datasets from the D4RL benchmark suite (\url{https://github.com/rail-berkeley/d4rl}) without any modification to the datasets. Specifically, we consider three MuJoCo locomotion tasks, namely, \texttt{halfcheetah}, \texttt{hopper}, and \texttt{walker2d}. 
The goal of each task is to move forward as fast as possible, while keeping the control cost minimal. 
For each task, we consider four types of datasets:
\begin{itemize}[leftmargin=5.5mm, topsep=-0.5pt, itemsep=-0.5pt]
    \item \texttt{random}: contains random policy rollouts;
    \item \texttt{medium}: contains medium-level policy rollouts;
    \item \texttt{medium-replay}: contains all samples seen while training a medium-level agent from scratch;
    \item \texttt{medium-expert}: half of the dataset consists of medium-level policy rollouts and the other half consists of expert-level policy rollouts. 
\end{itemize}

{\bf Training details for offline RL. }
For training CQL agents, following \citet{kumar2020conservative}, we built CQL on top of the publicly available implementation of SAC (\url{https://github.com/vitchyr/rlkit}). As for network architecture, we use 2-layer multi-layer perceptrons (MLPs) for value and policy networks (except for \texttt{halfcheetah-medium-expert} task where we found that 3-layer MLPs is more effective for training CQL agents). 
For all locomotion experiments, following the setup in \citet{kumar2020conservative}, we trained offline RL agents for 1000 epochs without early stopping.

{\bf Training details for online RL. }
For all locomotion experiments, we report the fine-tuning performance during 250K timesteps for 4 random seeds.
For our method, we used Adam optimizer \citep{kingma2014adam} with policy learning rates chosen from $\{3\mathrm{e}{-4}, 3\mathrm{e}{-5}, 5\mathrm{e}{-6}\}$, and value learning rate of $3\mathrm{e}{-4}$. We used ensemble size $N=5$
We found that taking 5 $\times$ training steps (5000 as opposed to the usual 1000) after collecting the first 1000 online samples slightly improves the fine-tuning performance.
After that, we trained for 1000 training steps every time 1000 additional samples were collected.

{\bf Training details for balanced replay.}
\label{supp:balanced_replay_setup}
For training the density ratio estimation network $w_{\psi}(s,a)$, which was fixed to be a 2-layer MLP for all our experiments, we used batch size 256 (i.e., 256 offline samples and 256 online samples), and learning rate $3\mathrm{e}{-4}$ for all locomotion experiments. When calculating the priority values, we applied self-normalization:
\begin{align}
\label{eq:selfnorm}
\widetilde{w}_{\psi}(x) = \frac{w_{\psi}(x)^{1/T}}{\mathbb{E}_{x\sim P}[w_{\psi}(x)^{1/T}]},
\end{align}
where $x$ and $P$ denote $(s,a)$ and $\mathcal{B}^{\tt{off}}$, respectively, and $T$ is the temperature hyperparameter. For all locomotion experiments, we used $T=5.0$. 
This stabilizes training by producing stable density ratios across random minibatches.

Before fine-tuning starts, we first add the offline samples to the prioritized replay buffer with default priority value $1.0$. Then, we set the default priority value to be high, in order to make sure online samples are seen enough for RL updates. In particular, letting $M$ denote the size of the offline buffer, we set the default priority value to be such that the initial 1000 online samples gathered would have probability $\rho$ of being seen, where $\rho$ is a hyperparameter, i.e., priority value of $ P_{0} := \frac{M}{1000} \cdot \frac{\rho}{1-\rho}$. We used $\rho \in \{0.5, 0.75\}$.

Note that after being seen once for an RL update, the given sample's priority value is appropriately updated to be $\tilde{w}_{\psi}(s,a)$. 
Afterwards, the default priority value is updated to be the maximum priority value seen during fine-tuning, as in \citep{schaul2015prioritized}. See Section~\ref{supp:pseudocode} for a pseudocode. 

{\bf Baselines. }
For AWAC \citep{nair2021awac}, we use the publicly available implementation from the authors (\url{https://github.com/vitchyr/rlkit/tree/master/examples/awac}) without any modification to hyperparameters and architectures.
For training BCQ \citep{fujimoto2019off}, we use the publicly available implementation from \url{https://github.com/rail-berkeley/d4rl_evaluations/tree/master/bcq}. For the online setup (BCQ-ft), we used additional online samples for training.
For SAC \citep{haarnoja2018soft}, we used the implementation from rlkit (\url{https://github.com/vitchyr/rlkit}) with default hyperparameters.

\newpage
\subsection{Robotic Manipulation Tasks}
\label{supp:manipulation_setup}

\begin{figure*}[ht!]
\centering
\hfill
\includegraphics[width=0.28\textwidth]{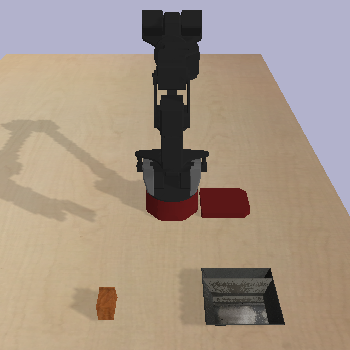}
\hfill
\includegraphics[width=0.28\textwidth]{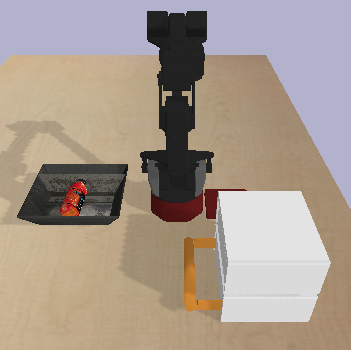}
\hfill
\includegraphics[width=0.28\textwidth]{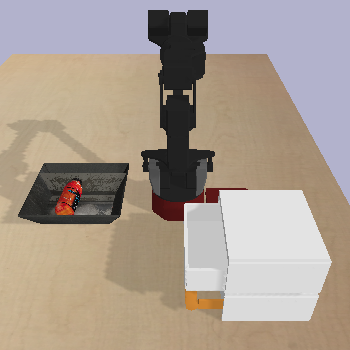}
\hfill
\caption{Manipulation tasks. (Left) \texttt{pick-place}, (Middle) \texttt{grasp-closed-drawer}, (Right) \texttt{grasp-blocked-drawer}.}
\label{supp:manip_task_fig}
\end{figure*}

\textbf{Task detail \& offline dataset.} 
We consider three sparse-reward pixel-based manipulation tasks from \citet{singh2020cog} (see Figure~\ref{supp:manip_task_fig}):
\begin{itemize}[leftmargin=5.5mm, topsep=-0.5pt, itemsep=-0.5pt]
\item \texttt{pick-place}: pick an object and put it in the tray. 
\item \texttt{grasp-closed-drawer}: grasp an object in the initially closed bottom drawer;  
\item \texttt{grasp-blocked-drawer}: grasp an object in the initially closed bottom drawer, where the initially open top drawer blocks the handle for the bottom drawer. 
\end{itemize}
Episode lengths for the tasks are 40, 50, 80, respectively. 

For each task, the offline dataset provided by the authors of \citet{singh2020cog} (\url{https://github.com/avisingh599/cog}) consists of highly structured transitions. For example, for \texttt{pick-place}, the dataset contains grasp attempts from a randomized scripted policy and place attempts from a randomized scripted policy; for \texttt{grasp-closed-drawer} and \texttt{grasp-blocked-drawer}, dataset contains various behaviors such as opening and closing both drawers, grasping object in the scene, and placing objects at random places.
For a more detailed description, refer to \citet{singh2020cog}. 
Meanwhile, it is rarely the case that logged data `in the wild' are generated by such structured, directed policies only. 
To consider a more realistic setup, we replace half of the original dataset with 100K uniform random policy rollouts. 
We remark that uniform random policy rollouts are commonly used in robotic tasks \citep{finn2017deep,ebert2018visual}.

{\bf Training details for offline RL. }
We used the official implementation by the authors of \citet{singh2020cog} (\url{https://github.com/avisingh599/cog}) for training offline CQL agents on the modified datasets, with default hyperparameters.  
Specifically, the value network is parametrized by three convolutional layers and two pooling layers in-between, followed by fully-connected layers of hidden sizes (1024, 512, 256) with ReLU activations. 
Kernels of size 3 $\times$ 3 with 16 channels are used for all three convolutional layers.
Maxpooling is applied after each of the first two convolutional layers, with stride 2. 
The policy network is identical to the value network, except it does not take action as an input. 
Policy learning rate and value learning rate are $1\mathrm{e}{-4}$ and $3\mathrm{e}{-4}$, respectively. 
For each task, we trained the agent for 500K train steps, then picked the best performing checkpoint in terms of average performance across seeds.

{\bf Training details for online RL.}
For our main results, we report fine-tuning performance during 1K episodes for 8 random seeds. For our method, we used Adam optimizer \citep{kingma2014adam} with policy learning rate $3\mathrm{e}{-5}$ and value learning rate $1\mathrm{e}{-4}$. Again, we applied 5 $\times$ training steps after collecting the first batch of samples. 
Architecture for density ratio estimation network $w_{\psi}$ is identical to the value network, except it has fully-connected layers with hidden sizes (256, 128). 
We used learning rate $3\mathrm{e}{-4}$ for $w_{\psi}$ and temperature $T=2.5$ for balanced replay in all manipulation tasks. We used ensemble size $N=4$.

{\bf Baseline. }
For CQL-online, we used the official implementation by the authors of \citet{singh2020cog} (\url{https://github.com/avisingh599/cog}). Specifically, the same CQL regularization is applied during online fine-tuning, while using online samples exclusively for updates. Learning rates are the same as in offline training.

\newpage
\section{Additional Experiment}
\subsection{Locomotion}
\label{supp:additional_exp_locomotion}

\begin{figure*}[ht!]
\vspace{-0.1in}
\centering
\subfloat{%
  \includegraphics[width=0.8\textwidth]{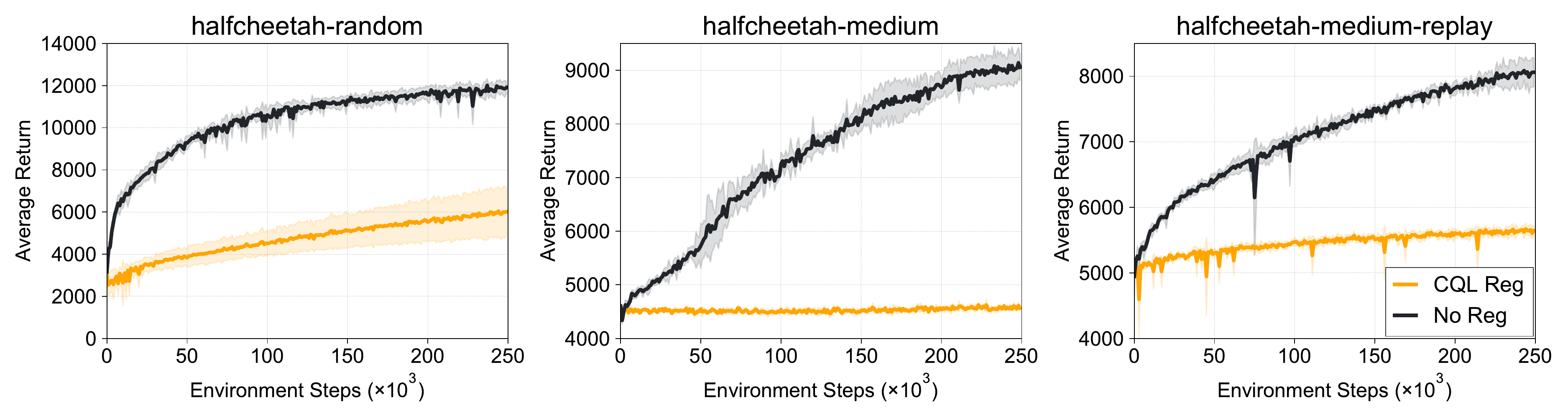}
}
\vspace{-0.05in}
\caption{
Fine-tuning performance with CQL regularization for \texttt{halfcheetah} tasks. The solid lines and shaded regions represent mean and standard deviation, respectively, across four runs.
}
\vspace{-0.15in}
\label{supp:cql_reg}
\end{figure*}

\begin{figure*}[ht!]
\vspace{-0.1in}
\centering
\subfloat{%
  \includegraphics[width=0.8\textwidth]{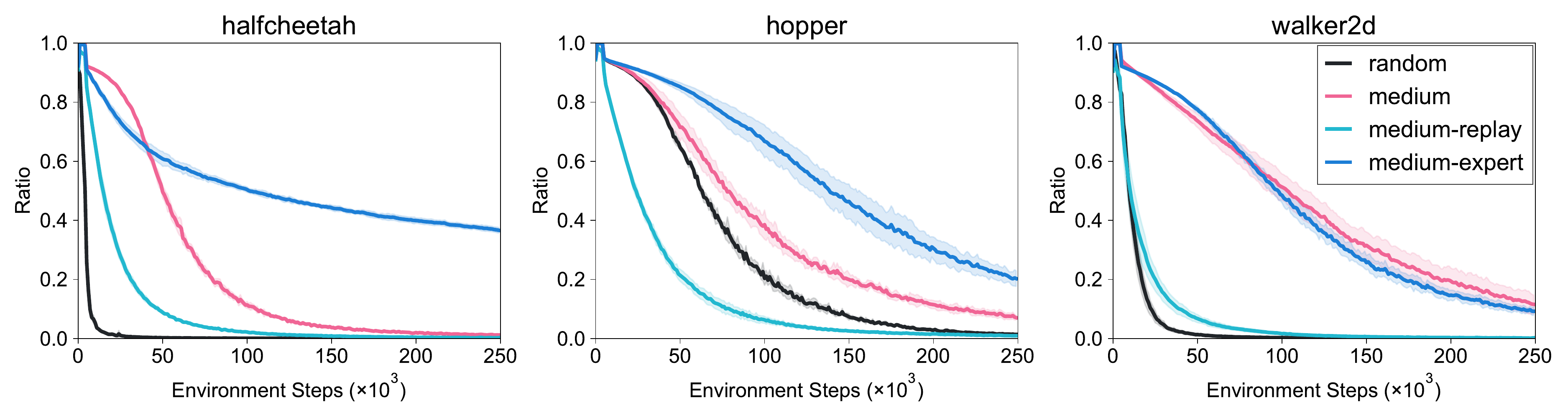}
}
\vspace{-0.05in}
\caption{
Proportion of offline samples used for updates during fine-tuning. The solid lines and shaded regions represent mean and standard deviation, respectively, across four runs. 
}
\vspace{-0.15in}
\label{supp:locomotion_buffer_analysis}
\end{figure*}

\begin{figure*}[ht!]
\vspace{-0.1in}
\centering
\subfloat{%
  \includegraphics[width=0.8\textwidth]{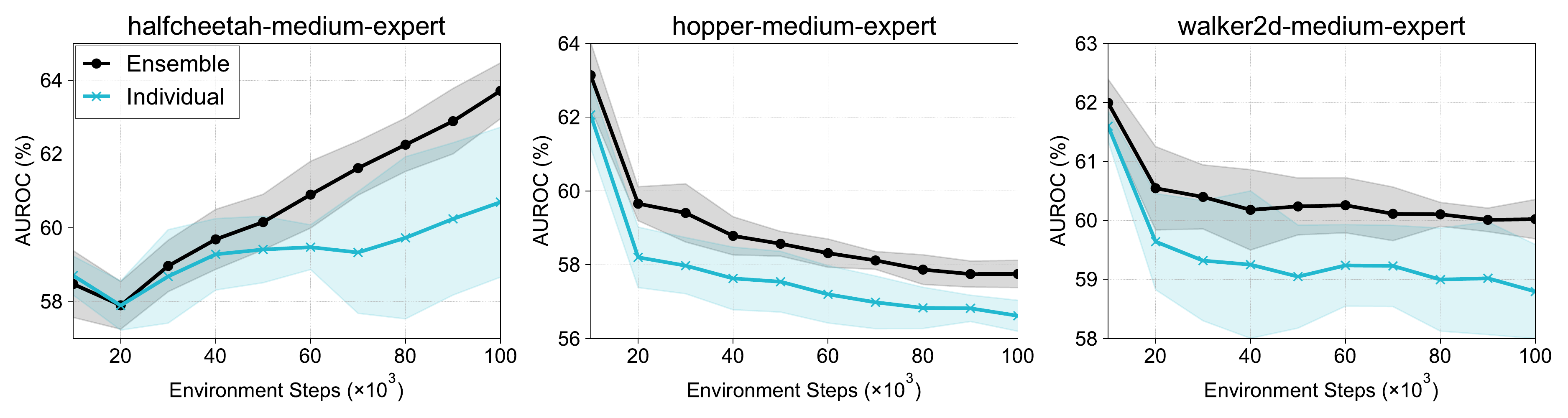}
}
\vspace{-0.05in}
\caption{
AUROC (\%) over the course of fine-tuning on \texttt{medium-expert} tasks, where the Q-function is interpreted as a binary classifier that classifies a given state-action pair $(s,a)$ as either a seen pair $(s, a_{\tt{seen}})$ or an unseen pair $(s,a_{\tt{uniform}})$, for a state $s$ encountered online. The solid lines and shaded regions represent mean and standard deviation, respectively, across four runs. 
}
\vspace{-0.05in}
\label{supp:ensemble_analysis}
\end{figure*}

{\bf Effects of CQL Regularization on Fine-tuning.}
Here, we show that removing the regularization term from the CQL objective $\mathcal{L}^{\tt{CQL}}_{\tt{critic}}$ is crucial for fast fine-tuning.
After training an offline RL agent via CQL (\ref{eq:cql}, \ref{eq:sac_actor}), we fine-tuned the agent via CQL objective (\ref{eq:cql}) (denoted CQL Reg), and via SAC objective (\ref{eq:sac_critic}) (denoted No reg), respectively. 
For both methods, balanced replay was applied. 
As shown in Figure~\ref{supp:cql_reg}, fine-tuning with CQL regularization prevents the agent from improving via online interaction, for pessimistic updates during fine-tuning (as opposed to pessimistic \textit{initialization}) essentially results in failure to explore. 
We remark that removing the regularization term may sometimes result in initial instability when working with narrowly distributed datasets, e.g., medium-expert, as seen in Figure~\ref{fig:performance}. For applications that significantly favor safety over rapid fine-tuning, more conservative approaches such as deployment-efficient RL \citep{matsushima2020deployment} may be more appropriate.

{\bf Balanced replay analysis.}
We provide buffer analysis as in Figure~\ref{fig:analysis_buffer} for all locomotion tasks. As seen in Figure~\ref{supp:locomotion_buffer_analysis}, trends are similar across all environments, i.e., balanced replay automatically decides whether offline samples will be used or not depending their relevance to the current agent.

{\bf Q-ensemble analysis.}
\label{sup:ensemble_analysis}
We provide further details and experiments for Q-ensemble analysis from Figure~\ref{fig:abl_discriminative}. 
Letting $\mathcal{D}^{\tt{real}}_{T} := \{(s_{i},a_{i})\}_{i=1}^{T}$ be the samples collected online up until timestep $T$, 
we construct a corresponding fake dataset by replacing the actions in $\mathcal{D}^{\tt{real}}_{T}$ by uniform random actions, i.e., $\mathcal{D}^{\tt{fake}}_{T} := \{(s_{i},a_{\tt{unif}})\}_{i=1}^{T}$. 
Note that to provide enough coverage of the action space, we sampled $50$ random actions for each real state-action pair considered. 
Then, interpreting $Q(s,a)$ as the confidence value for classifying real and fake transitions, we measure the area under ROC (AUROC) curve values over the course of fine-tuning. 
As seen in Figure~\ref{supp:ensemble_analysis}, Q-ensemble demonstrates superior discriminative ability, which prevents distribution shift and leads to stable fine-tuning. 

\newpage
\begin{figure*}[t!]
\centering
\includegraphics[width=0.95\textwidth]{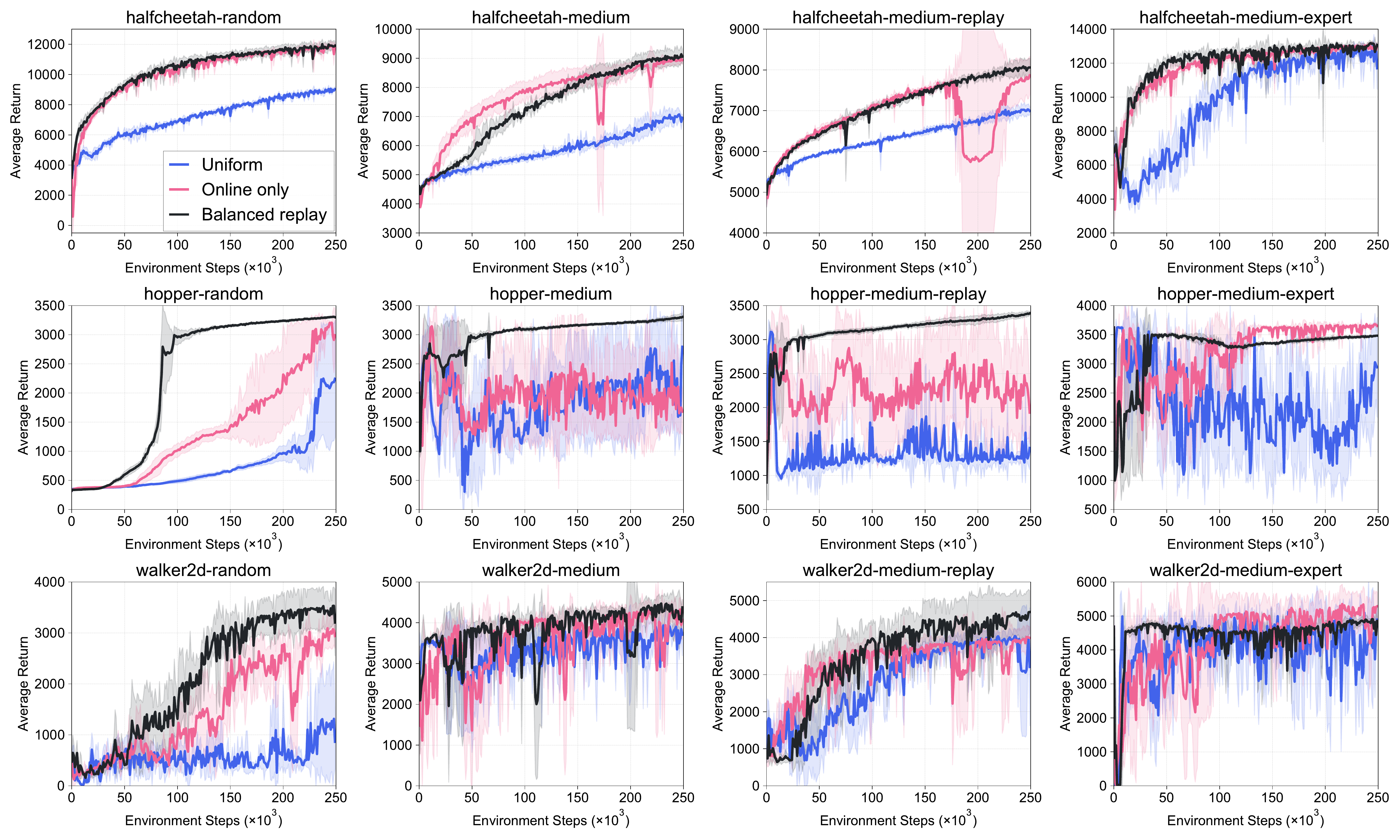}
\caption{Ablation on balanced replay. The solid lines and shaded regions represent mean and standard deviation, respectively, across four runs.
}
\label{supp:abl_buffer}
\end{figure*}

\begin{figure*}[t!]
\centering
\includegraphics[width=0.95\textwidth]{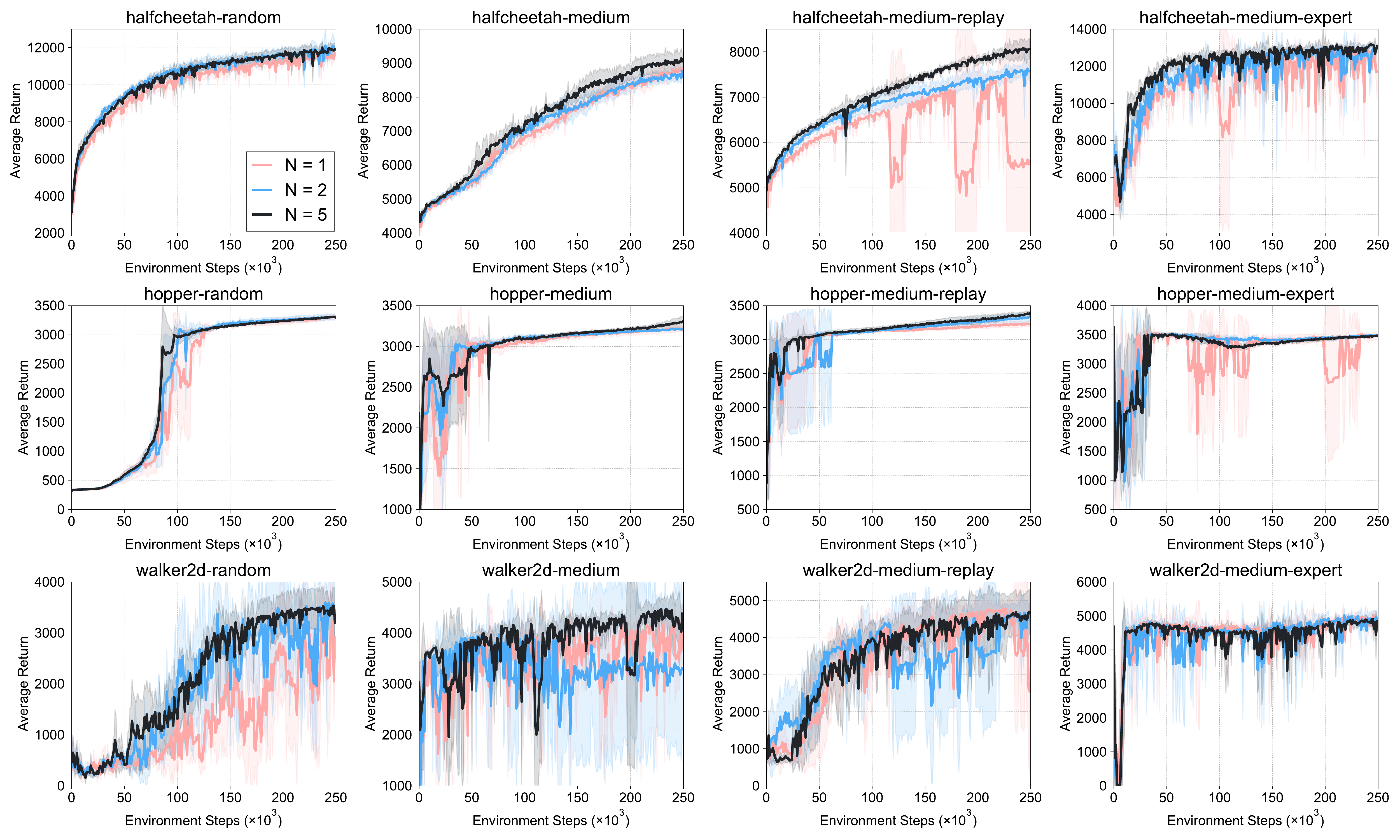}
\caption{Fine-tuning performance with varying ensemble sizes. The solid lines and shaded regions represent mean and standard deviation, respectively, across four runs.
}
\label{supp:abl_size}
\end{figure*}

{\bf Ablation on balanced replay.}
We provide ablation study on balanced replay as in Figure~\ref{fig:abl_experience_replay} for all tasks (Figure~\ref{supp:abl_buffer}). Balanced replay provides significant performance gain compared to the other two sampling strategies: Online only, and Uniform. The benefit is especially pronounced in the more complex environment of \texttt{hopper} and \texttt{walker2d}.  

{\bf Ablation on ensemble size.}
We provide learning curves for all tasks with varying ensemble size $N \in \{1, 2, 5\}$. As seen in Figure\ref{supp:abl_size}, performance improves as $N$ grows. However, due to the trade-off between computational overhead and performance, we opted for $N=5$ for our method.

\newpage

\begin{figure*}[t!]
\centering
\includegraphics[width=0.95\textwidth]{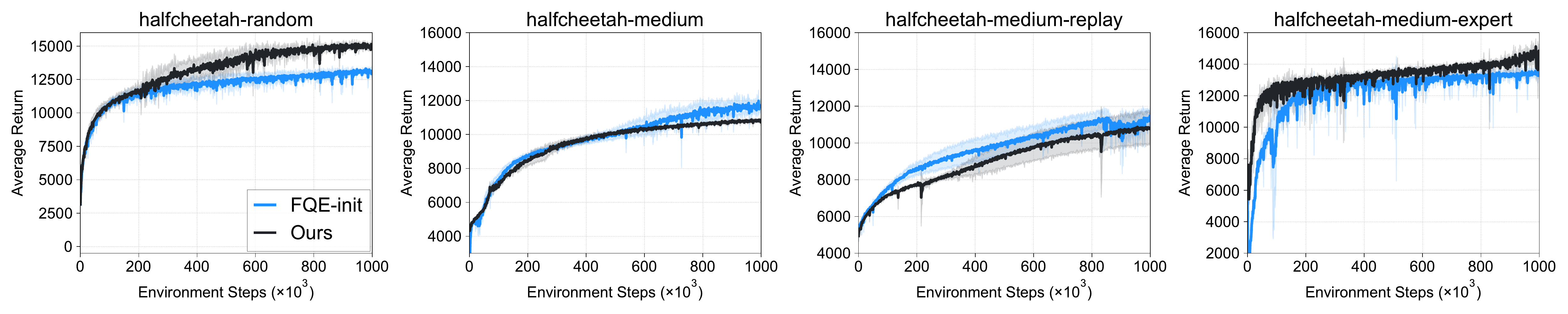}
\caption{Asymptotic performance of our method (pessimistic Q-function) and FQE-init that fine-tunes a neutral Q-function. Ensemble and balanced replay were applied to both methods. The solid lines and shaded regions represent mean and standard deviation, respectively, across four runs.}
\label{supp:exploration_asymptotic}
\end{figure*}

\begin{figure*}[t!]
\vspace{-0.15in}
\centering
\includegraphics[width=0.95\textwidth]{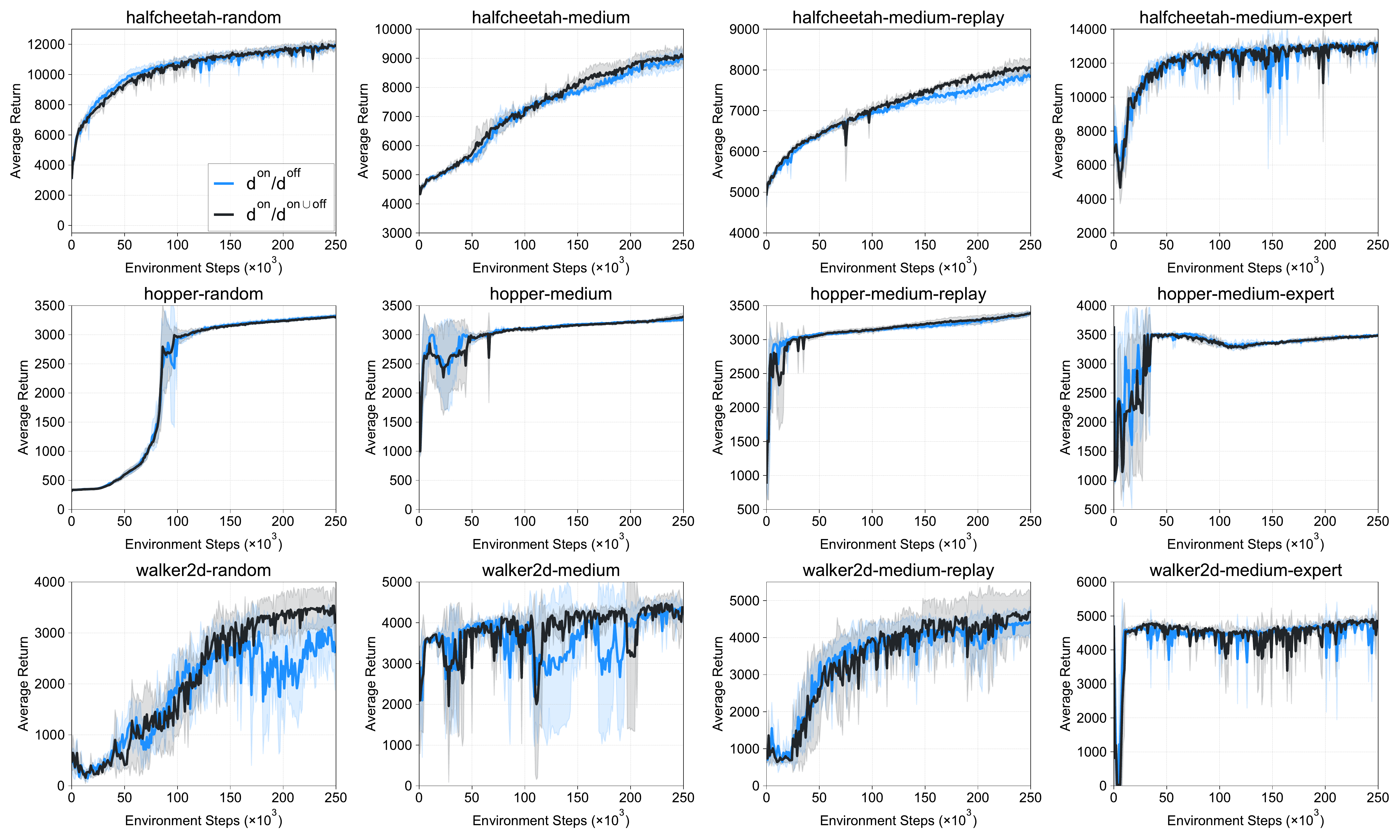}
\caption{An alternative density ratio ($d^{\tt{on}}/d^{\tt{on \cup off}}$) for balanced replay. The solid lines and shaded regions represent mean and standard deviation, respectively, across four runs.}
\label{supp:density_ratio}
\vspace{-0.25in}
\end{figure*}

{\bf Dataset composition and performance.}
In our experiments, we observed that dataset composition may be related to the fine-tuning performance in a counterintuitive way. In order to investigate this, we additionally ran our method on halfcheetah tasks for 1 million steps. As shown in Figure~\ref{supp:exploration_asymptotic}, our method performed best on random dataset, reaching 15k average return, while reaching only 12k for medium and medium-replay datasets. We conjecture this is because the agent sees the most diverse set of samples when trained from scratch. This means that Q-function can extrapolate better, and the agent can explore more efficiently as a result. While the medium-replay dataset also contains diverse samples, it asymptotically performs worst, possibly because it contains significantly less data compared to the other datasets -- random and medium contain 1 million transitions, respectively, and medium-expert contains 2 million transitions. 

We note that failure to explore properly is not necessarily due to the pessimism in Q-function, but is rather a legacy of the dataset composition. To show this, we fine-tuned an ensemble of FQE-initialized agents (neutral Q-function) with balanced replay (see Section~\ref{subsec:qfunction} and Section~\ref{supp:concept} for details) then compared it with our method (pessimistic Q-function), as in Section~\ref{subsec:qfunction}. As shown in Figure~\ref{supp:exploration_asymptotic}, we observe that the two methods show more or less similar asymptotic performances with the exception of halfcheetah-random, where our method noticeably outperforms FQE-init. This shows that initial pessimism does not harm the asymptotic online performance, while preventing initial performance degradation in many cases (meanwhile, FQE-init suffers from performance degradation for medium and medium-expert tasks, a result in line with the observation from Section~\ref{subsec:qfunction}).

{\bf Exploring different density ratio schemes.}
We also additionally experimented with density ratio $w(s,a) := d^{\tt{on}}(s,a)/ d^{\tt{off \cup \texttt{on}}}(s,a)$ instead of $w(s,a) := d^{\tt{on}}(s,a)/ d^{\tt{off}}(s,a)$ for balanced replay. The hyperparameters were set identically to the original setup.
As seen in Figure~\ref{supp:density_ratio}, we observed similar performance across most setups. However, the alternative scheme may have the potential benefit of trending towards $1.0$ asymptotically.

\newpage
\subsection{Manipulation}

\begin{figure*}[t!]
\centering
\hfill
\subfloat{
  \includegraphics[width=0.95\textwidth]{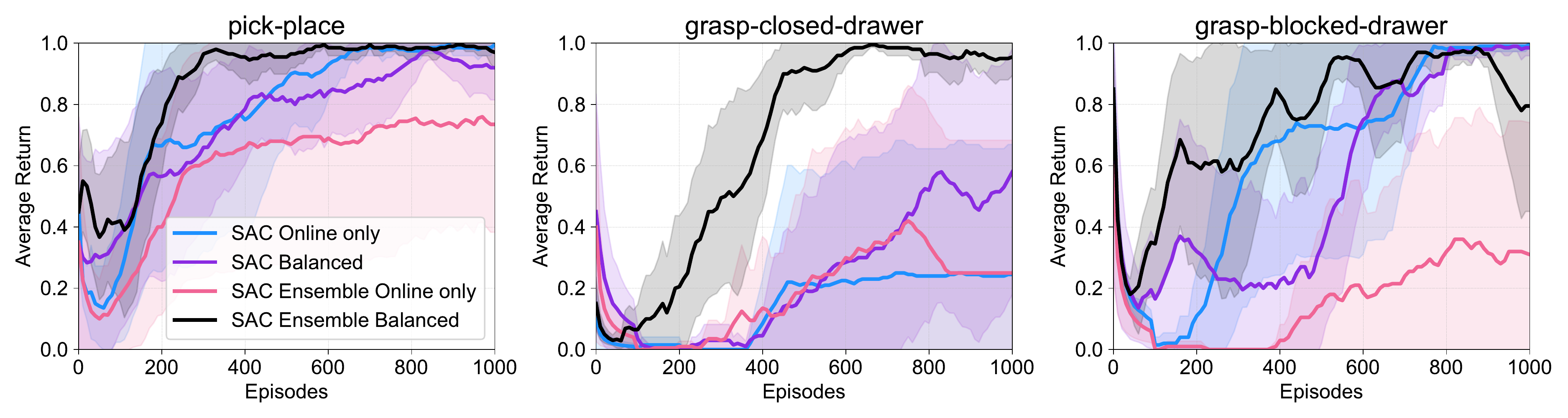}
}
\hfill
\vspace{-0.05in}
\caption{
Ablation study with various ablated versions of our method, but with SAC objective. 
The solid lines and shaded regions represent mean and standard deviation, respectively, across four runs.
}
\vspace{-0.2in}
\label{supp:manipulation_ablation_sac}
\end{figure*}

\begin{figure*}[t!]
\centering
\hfill
\subfloat{
  \includegraphics[width=0.95\textwidth]{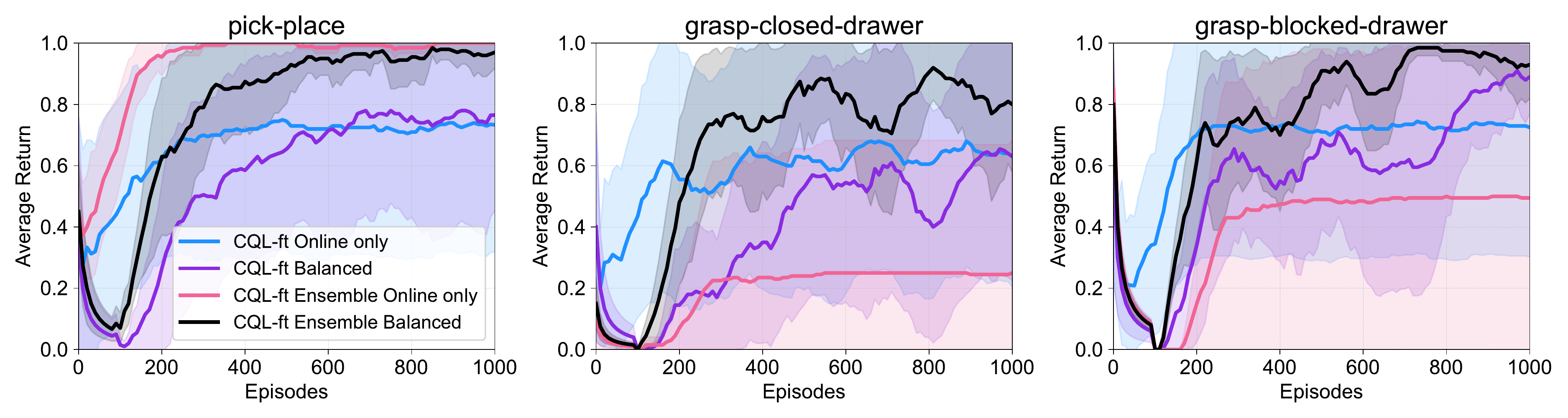}
}
\hfill
\vspace{-0.05in}
\caption{
Ablation study with various ablated versions of our method, but with CQL objective. 
The solid lines and shaded regions represent mean and standard deviation, respectively, across four runs.
}
\vspace{-0.15in}
\label{supp:manipulation_ablation_cql}
\end{figure*}

\begin{figure*}[t!]
\centering
\hfill
\subfloat{%
  \includegraphics[width=0.95\textwidth]{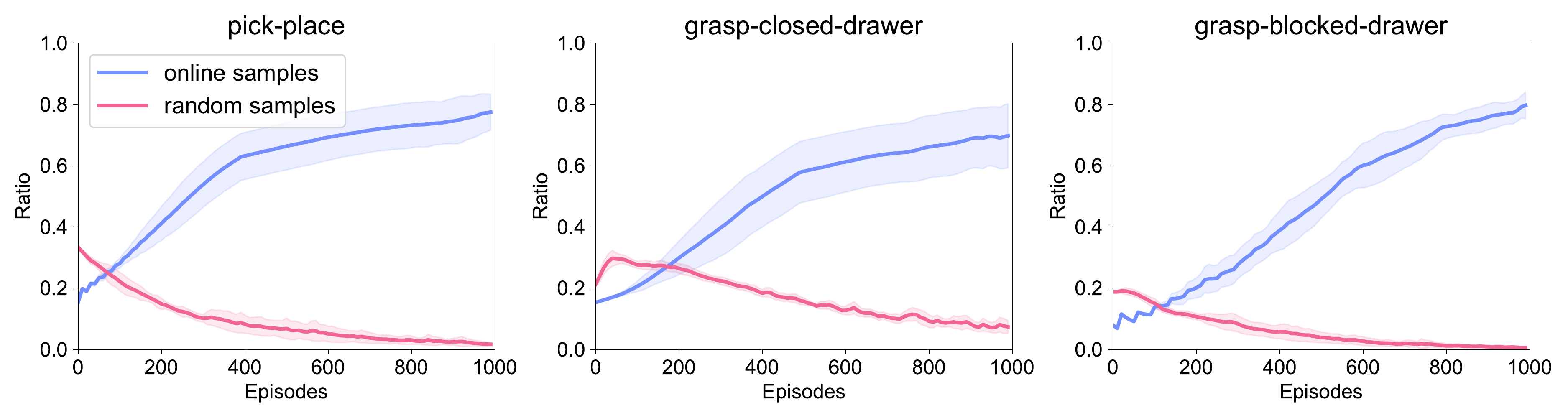}
}
\hfill
\vspace{-0.05in}
\caption{
Proportion of random data used for during fine-tuning decreases over time.  
The solid lines and shaded regions represent mean and standard deviation, respectively, across four runs.
}
\vspace{-0.1in}
\label{supp:manipulation_buffer}
\end{figure*}

{\bf Ablation studies.} 
We provide ablation studies of our method for the robotic manipulation tasks (Figure~\ref{supp:manipulation_ablation_sac}). Starting from offline CQL agents, we consider:
\begin{itemize}[leftmargin=5.5mm, topsep=-0.5pt, itemsep=-0.5pt]
    \item SAC Online only: Fine-tune with SAC, with online samples only;
    \item SAC Balanced: Fine-tune with SAC, with balanced replay;
    \item SAC Ensemble Online only: Fine-tune an ensemble agent with SAC, with online samples only;
    \item SAC Ensemble Balanced: Ours, i.e., fine-tune an ensemble agent with SAC and balanced replay.
\end{itemize}
Furthermore, we additionally experimented with CQL objectives instead of SAC objectives, and provide the results (Figure~\ref{supp:manipulation_ablation_cql}).  

In most cases, both ensemble and balanced replay were essential for good performance. First, due to reward sparsity, there were individual offline agents that, while producing meaningful trajectories that resemble some trajectories in the offline dataset, were nonetheless not good enough to receive reward signals online. In this case, the ensemble agent becomes more robust than these individual agents, and was more likely to see enough reward signals for fine-tuning. Second, even with ensemble, updating with online samples only resulted in severe distribution shift, and balanced replay was essential to facilitate fine-tuning. However, note that it only took our method about 200 episodes (around 8k to 16k environment steps) to recover the initial score, and that our method achieves best asymptotic scores both with SAC and CQL objectives.

{\bf Balanced replay analysis.} 
We provide analysis for balanced replay in all manipulation tasks considered.  
As seen in Figure~\ref{supp:manipulation_buffer}, without any privileged information, balanced replay automatically selects relevant offline samples for updates, while filtering out task-irrelevant, random data as fine-tuning proceeds.

\end{document}